# A Theory of Cross-Validation Error

Peter Turney
Knowledge Systems Laboratory
Institute for Information Technology
National Research Council Canada
Ottawa, Ontario, Canada
K1A 0R6

613-993-8564
peter@ai.iit.nrc.ca

**Running Head:** A Theory of Cross-Validation Error

# A Theory of Cross-Validation Error

## Abstract

This paper presents a theory of error in cross-validation testing of algorithms for predicting real-valued attributes. The theory justifies the claim that predicting real-valued attributes requires balancing the conflicting demands of simplicity and accuracy. Furthermore, the theory indicates precisely how these conflicting demands must be balanced, in order to minimize cross-validation error. A general theory is presented, then it is developed in detail for linear regression and instance-based learning.

## 1 Introduction

This paper is concerned with cross-validation testing of algorithms that perform supervised learning from examples. It is assumed that each example is described by a set of attribute-value pairs. The learning task is to predict the value of one of the attributes, given the values of the remaining attributes. It is assumed that the attributes, both the predictor attributes and the attribute to be predicted, range over real numbers, and that each example is described by the same set of attributes.

Let us suppose that the examples have $r + 1$ attributes. We may represent the examples as points in $r + 1$ dimensional real space $\Re^{r+1}$. The task is to learn a function that maps from $\Re^r$ to $\Re$, where $\Re^r$ is the space of the predictor attributes and $\Re$ is the space of the prediction.

In cross-validation testing, a teacher gives the learning algorithm (the *student*) a set of training examples (hence *supervised* learning), consisting of points in $\Re^{r+1}$. The algorithm uses these data to form a model. The teacher then gives the algorithm a set of testing examples, consisting of points in $\Re^r$. The value of the attribute that is to be predicted is hidden from the student, but known to the teacher. The student uses its model to calculate the value of the hidden attribute. The student is scored by the difference between the predictions it makes and the actual values for the hidden attribute in the testing examples. This difference is the error of the algorithm in cross-validation testing.

Section 2 presents a theory of error in cross-validation testing. It is shown that we may think of cross-validation error as having two components. The first component is the error of the algorithm on the training set. The error on the training set is often taken to be a measure of the *accuracy* of the algorithm (Draper and Smith, 1981). The second component is the *instability* of the algorithm (Turney, 1990). The instability of the algorithm is the sensitivity of the algorithm to noise in the data. Instability is closely related to our intuitive notion of complexity (Turney, 1990). Complex models tend to be unstable and simple models tend to be stable.

It is proven that cross-validation error is limited by the sum of the training set error and the instability. This theorem is justification for the intuition that good models should be both simple and accurate. It is assumed that a good model is a model that minimizes cross-validation error.

After examining cross-validation error in general, Section 3 looks at the accuracy and stability of linear regression (Draper and Smith, 1981), and Section 4 considers instance-based learning (Kibler *et al*., 1989; Aha *et al.*, 1991). In both cases, it turns out that there is a conflict between accuracy and stability. When we try to maximize accuracy (minimize error on the training set), we find that stability tends to decrease. When we try to maximize stability (maximize resistance to noise), we find that accuracy tends to decrease. If our goal is to minimize cross-validation error, then we must find a balance between the conflicting demands of accuracy and stability. This balance can be found by minimizing the sum of the error on the training set and the instability.

The theoretical discussion is followed by Section 5, which considers the practical application of the theory. Section 5 presents techniques for estimating accuracy and stability. The theory is then applied to an empirical comparison of instance-based learning and linear regression (Kibler *et al*. 1989). It is concluded that the domain will determine whether instance-based learning or linear regression is superior.

Section 6 presents an example of fitting data with linear regression and instance-based learning. It is argued that the formal concept of stability captures an important aspect of our intuitive notion of simplicity. This claim cannot be proven, since it involves intuition. The claim is supported with the example.

Section 7 compares this theory with the work in Akaike Information Criterion (AIC) statistics (Sakamoto *et al.*, 1986). There are interesting similarities and differences between the two approaches.

Finally, Section 8 considers future work. One weakness of the theory that needs to be addressed is cross-validation with a testing set that requires interpolation and extrapolation. This weakness implies that the theory may underestimate cross-validation error. Another area for future work is extending these results to techniques other than linear regression and instance-based learning. The theory should also be extended to handle predicting symbolic attributes, in addition to real-valued attributes.

# 2 Cross-Validation Error

Suppose we have an experimental set-up that can be represented as a black box with $r$ inputs and one output, where the inputs and output can be represented by real numbers. Let us imagine that the black box has a deterministic aspect and a random aspect. We may use the function $f$, where $f$ maps from $\Re^r$ to $\Re$, to represent the deterministic aspect of the black box. We may use the random variable $z$ to represent the random aspect of the black box. We may assume that $z$ is a sample from a standardized distribution, with mean 0 and variance 1. We may use the constant $\sigma$ to scale the variance of the random variable $z$ to the appropriate level. Let the vector $\vec{v}$ represent the inputs to the black box for a single experiment, where:

$$\vec{v} = \begin{bmatrix} v_1 & \dots & v_r \end{bmatrix} \tag{1}$$

Let $y$ represent the output of the black box. Our model of the experiment is:

$$y = f(\vec{v}) + \sigma z \tag{2}$$

That is, the output of the black box is a deterministic function of the inputs, plus some random noise.

Suppose we perform $n$ experiments with the black box. For each experiment, we record the input values and the output value. Let us use the matrix $X$ to represent all of the inputs:

$$X = \begin{bmatrix} x_{1,1} & \dots & x_{1,r} \\ \dots & x_{i,j} & \dots \\ x_{n,1} & \dots & x_{n,r} \end{bmatrix} \tag{3}$$

Let the $i$-th row of the matrix $X$ be represented by the vector $\vec{v}_i$, where:

$$\vec{v}_i = \begin{bmatrix} x_{i,1} & \dots & x_{i,r} \end{bmatrix} \qquad i = 1, \dots, n \tag{4}$$

The vector $\vec{v}_i$ contains the values of the $r$ inputs for the $i$-th experiment. Let the $j$-th column of the matrix $X$ be represented by the vector $\vec{x}_j$, where:

$$\vec{x}_j = \begin{bmatrix} x_{1,j} \\ \dots \\ x_{n,j} \end{bmatrix} \qquad j = 1, \dots, r \tag{5}$$

The vector $\vec{x}_j$ contains the values of the $j$-th input for the $n$ experiments. Let the $n$ outputs be represented by the vector $\vec{y}$, where:

$$\vec{y} = \begin{bmatrix} y_1 \\ \dots \\ y_n \end{bmatrix} \tag{6}$$

The scalar $y_i$ is the output of the black box for the $i$-th experiment.

The function $f$ can be extended to a vector function $\vec{f}(X)$, where:

$$\vec{f}(X) = \begin{bmatrix} f(\vec{v}_1) \\ \dots \\ f(\vec{v}_n) \end{bmatrix} \tag{7}$$

Our model for the $n$ experiments is:

$$\vec{y} = \vec{f}(X) + \sigma\vec{z} \tag{8}$$

The vector $\vec{z}$ is a sequence of $n$ independent samples from a standardized distribution:

$$\vec{z} = \begin{bmatrix} z_1 \\ \dots \\ z_n \end{bmatrix} \tag{9}$$

Imagine an infinite sequence of repetitions of the whole set of $n$ experiments, with $X$ held constant:

$$\vec{y}_1 = \vec{f}(X) + \sigma\vec{z}_1$$
$$\vec{y}_2 = \vec{f}(X) + \sigma\vec{z}_2 \tag{10}$$
$$\ldots$$

With each repetition $k$ of the $n$ experiments, the $n$ outputs $\vec{y}_k$ change, because the random noise $\vec{z}_k$ has changed:

$$\vec{y}_k = \begin{bmatrix} y_{k,1} \\ \ldots \\ y_{k,n} \end{bmatrix} \qquad \vec{z}_k = \begin{bmatrix} z_{k,1} \\ \ldots \\ z_{k,n} \end{bmatrix} \tag{11}$$

Consider the average output $\vec{y}_a$ of the first $m$ repetitions:

$$\vec{y}_a = \begin{bmatrix} y_{a,1} \\ \ldots \\ y_{a,n} \end{bmatrix} \qquad y_{a,i} = \frac{1}{m}\sum_{j=1}^{m} y_{j,i} \tag{12}$$

By the Weak Law of Large Numbers (Fraser, 1976), as $m \to \infty$, $\vec{y}_a$ converges to $\vec{f}(X)$. This follows from the fact that the mean of the noise $\vec{z}_k$ is zero. Thus, if we could actually repeat the whole set of $n$ experiments indefinitely many times, then we could find the true value of the deterministic aspect $f$ of the black box, for input $X$, with arbitrary accuracy.

Suppose that we are trying to develop a model of $f$. Let us write $m(\vec{v}|X,\vec{y})$ to represent the prediction that the model makes for $f(\vec{v})$, when the model is based on the data $X$ and $\vec{y}$. Let us apply the model to the data $X$ and $\vec{y}$ on which the model is based:

$$\vec{m}(X|X,\vec{y}) = \begin{bmatrix} m(\vec{v}_1|X,\vec{y}) \\ \ldots \\ m(\vec{v}_n|X,\vec{y}) \end{bmatrix} \tag{13}$$

Thus $\vec{m}(X|X,\vec{y})$ is the model's prediction for $\vec{f}(X)$.

Consider the first two sets of experiments in the above sequence (10):

$$\begin{aligned} \vec{y}_1 &= \grave{f}(X) + \sigma \vec{z}_1 \\ \vec{y}_2 &= \grave{f}(X) + \sigma \vec{z}_2 \end{aligned} \tag{14}$$

Suppose that the data $(X, \vec{y}_1)$ are the training set and the data $(X, \vec{y}_2)$ are the testing set in cross-validation testing of the model *m*. The error on the training set $\vec{e}_t$ is:

$$\vec{e}_t = \vec{m}(X | X, \vec{y}_1) - \vec{y}_1 = \begin{bmatrix} e_{t,1} \\ \dots \\ e_{t,n} \end{bmatrix} \tag{15}$$

The error on the testing set, the cross-validation error $\vec{e}_c$, is:

$$\vec{e}_c = \vec{m}(X | X, \vec{y}_1) - \vec{y}_2 = \begin{bmatrix} e_{c,1} \\ \dots \\ e_{c,n} \end{bmatrix} \tag{16}$$

Let us assume that our goal is to minimize the expected length of the cross-validation error vector:

$$E(\| \vec{e}_c \|) = E\left(\sqrt{\vec{e}_c^{\mathrm{T}} \vec{e}_c}\right) = E\left(\sqrt{\sum_{i=1}^{n} e_{c,i}^2}\right) \tag{17}$$

T is the matrix transpose operation, $\| \dots \|$ is vector length, and $E(\dots)$ is the expectation operator of probability theory (Fraser, 1976). If $t(x)$ is a function of a random variable *x*, where *x* is a sample from a probability distribution with density $p(x)$, then the expected value of $t(x)$ is defined as follows (Fraser, 1976):

$$E(t(x)) = \int_{-\infty}^{\infty} t(x) p(x) dx \tag{18}$$

The expected value of $t(x)$ is its mean or average value. Note that the integration in $E(\| \vec{e}_c \|)$ is over both $\vec{y}_1$ and $\vec{y}_2$.

It is assumed that the inputs *X* are the same in the training set and the testing set. This assumption is the main limitation of this theory of cross-validation error. The assumption is reasonable when we have a laboratory situation, where we can set the inputs to the black box to be whatever we want them to be. Otherwise — with data collected purely by observation, for example — the assumption may seem unreasonable.

The main reason for making the assumption is that it makes the mathematics simpler. If we have one set of inputs $X_1$ for the training set and another set of inputs $X_2$ for the testing set, then we can say very little about cross-validation error, unless we make some assumptions about *f*. If we assume that $X_1$ equals $X_2$, then we can prove some interesting results without making any assumptions about *f*. The assumption that $X_1$ equals $X_2$ is not onerous. Even outside of a laboratory, with enough data, we can select $X_2$ so that it closely approximates $X_1$.

We may expect a model to perform less well when the inputs on testing $X_2$ are significantly different from the inputs on training $X_1$. Therefore the main implication of the assumption is that we may be underestimating the cross-validation error.

A model $m$ of *f* typically has two possible interpretations, a causal interpretation and a predictive interpretation. For example, suppose that $y$ is the number of cigarettes that a person smokes in a week. In a predictive model, one could include lung cancer as an input in $m$, because it helps to predict $y$. In a causal model, one would not include lung cancer as an input, because lung cancer does not cause smoking. The work in this paper addresses predictive models, not causal models, although it may be applicable to causal models. The inputs $X$ may be any variables that are relevant for predicting $y$. The noise $z$ can represent all of the unmeasured causes of the output $y$ or an irreducible chance element in a nondeterministic world. [1]

The black box metaphor tends to suggest a causal connection between the inputs and the outputs, but the mathematics here deals only with prediction. The metaphor of a black box was chosen to make it seem reasonable that $X_1$ and $X_2$ are identical. However, as was discussed above, the results here are applicable even when the data are collected purely by observation, not experimentation.

There is another form of error $\grave{e}_s$, which we may call the *instability* of the model *m*:

$$\vec{e}_s = \vec{m}(X|X,\vec{y}_2) - \vec{m}(X|X,\vec{y}_1) = \begin{bmatrix} e_{s,1} \\ \dots \\ e_{s,n} \end{bmatrix} \tag{19}$$

Instability is a measure of how sensitive our modeling procedure is to noise in the data. It is the difference between the best fit of our model for the data $(X,\vec{y}_1)$ and the best fit for the data $(X,\vec{y}_2)$. If our modeling procedure resists noise (i.e. it is stable), then the two fits should be virtually the same, and thus $E(\|\vec{e}_s\|)$ should be small. If our modeling procedure is sensitive to noise (i.e. it is unstable), then $E(\|\vec{e}_s\|)$ should be large.

Now we are ready for the main result of this section:

**Theorem 1:** The expected size of the cross-validation error is less than or equal to the sum of the expected size of the training set error and the expected size of the instability:

$$E(\|\vec{e}_c\|) \le E(\|\vec{e}_t\|) + E(\|\vec{e}_s\|) \tag{20}$$

*Proof*: Let us introduce a new term:

$$\vec{e}_\omega = \vec{m}(X|X,\vec{y}_2) - \vec{y}_1 \tag{21}$$

By the symmetry of the training set and the testing set (14, 16, 21):

$$E(\|\vec{e}_c\|) = E(\|\vec{e}_\omega\|) \tag{22}$$

Note that (15, 19, 21):

$$\vec{e}_\omega = \vec{e}_t + \vec{e}_s \tag{23}$$

Therefore, by the triangle inequality:

$$\|\vec{e}_\omega\| = \|\vec{e}_t + \vec{e}_s\| \le \|\vec{e}_t\| + \|\vec{e}_s\| \tag{24}$$

Finally:

$$E(\|\vec{e}_c\|) = E(\|\vec{e}_t + \vec{e}_s\|) \le E(\|\vec{e}_t\| + \|\vec{e}_s\|) = E(\|\vec{e}_t\|) + E(\|\vec{e}_s\|) \tag{25}$$

$\square$.

We may interpret Theorem 1 as follows. The term $E(\|\vec{e}_t\|)$ is a measure of the accuracy of our model *m*. The term $E(\|\vec{e}_s\|)$ captures an aspect of our intuitive notion of the simplicity of *m*. It is often said that we should seek models that best balance the con-

flicting demands of accuracy and simplicity. As we shall see in the next two sections, there is some conflict between $E(\|\vec{e}_t\|)$ and $E(\|\vec{e}_s\|)$. When we attempt to minimize one of these terms, the other tends to maximize. If we minimize the sum $E(\|\vec{e}_t\|) + E(\|\vec{e}_s\|)$, then we can set an upper bound on the expected cross-validation error $E(\|\vec{e}_c\|)$. Thus Theorem 1 supports the traditional view that models should be both simple and accurate. Note that Theorem 1 requires no special assumptions about the form of either the deterministic aspect *f* of the black box or the random *z* aspect.

The following theorem sets a lower bound on cross-validation error:

**Theorem 2:** If *f* is known, then the best strategy to minimize the expected cross-validation error $E(\|\vec{e}_c\|)$ is to set the model equal to *f*:

$$\vec{m}(X|X,\vec{y}_i) = \vec{f}(X) \tag{26}$$

When the model is equal to *f*, we have:

$$E(\|\vec{e}_s\|) = 0 \tag{27}$$

$$E(\|\vec{e}_t\|) = E(\|\vec{e}_c\|) \tag{28}$$

$$E(\|\vec{e}_c\|^2) = \sigma^2 n \tag{29}$$

$$E(\|\vec{e}_c\|) \le \sigma\sqrt{n} \tag{30}$$

*Proof:* Suppose that we set the model equal to *f*:

$$\vec{m}(X|X,\vec{y}_i) = \vec{f}(X) \tag{31}$$

Consider the stability of the model:

$$\|\vec{e}_s\| = \|\vec{m}(X|X,\vec{y}_2) - \vec{m}(X|X,\vec{y}_1)\| = \|\vec{f}(X) - \vec{f}(X)\| = 0 \tag{32}$$

$$E(\|\vec{e}_s\|) = 0 \tag{33}$$

Since the model is fixed, it is perfectly stable. Let us introduce a new term:

$$\vec{e}_\omega = \vec{m}(X|X,\vec{y}_2) - \vec{y}_1 \tag{34}$$

We see that:

$$\vec{e}_\omega = \vec{m}(X|X,\vec{y}_1) - \vec{y}_1 = \vec{e}_t \tag{35}$$

Therefore:

$$E(\|\vec{e}_t\|) = E(\|\vec{e}_\omega\|) = E(\|\vec{e}_c\|) \tag{36}$$

In other words, when the model is equal to *f*, the expected magnitude of the cross-validation error equals the expected magnitude of the training set error. Furthermore, the expected magnitude of the cross-validation error is equal to the expected magnitude of the noise:

$$E(\|\vec{e}_c\|) = E(\|\vec{m}(X|X,\vec{y}_1) - \vec{y}_2\|) = E(\|\vec{f}(X) - \vec{y}_2\|) = E(\|\sigma\vec{z}_2\|) \quad (37)$$

Since $E(\|\vec{e}_c\|) = E(\|\sigma\vec{z}_2\|)$, it follows that the expected magnitude of the cross-validation error $E(\|\vec{e}_c\|)$ is minimal. Recall that $y = f(\vec{v}) + \sigma z$, by definition. Thus the output $y$ necessarily contains the random noise $z$. Since there is no way to predict the noise $\vec{z}_2$, there is no way to reduce $E(\|\vec{e}_c\|)$ below the level $E(\|\sigma\vec{z}_2\|)$. [2] This proves that the best strategy to minimize the expected cross-validation error $E(\|\vec{e}_c\|)$ is to set the model equal to *f*. Now, consider the squared length of the cross-validation error:

$$E(\|\vec{e}_c\|^2) = E(\|\sigma\vec{z}_2\|^2) = \sigma^2 E(\|\vec{z}_2\|^2) = \sigma^2 E\left(\sum_{i=1}^{n} z_{2,i}^2\right) \quad (38)$$

Since $\vec{z}_2$ is a sequence of independent samples from a standardized distribution (mean 0, variance 1), it follows that (Fraser, 1976):

$$E(\|\vec{e}_c\|^2) = \sigma^2\left(\sum_{i=1}^{n} E(z_{2,i}^2)\right) = \sigma^2\left(\sum_{i=1}^{n} 1\right) = \sigma^2 n \quad (39)$$

Finally, applying Lemma 1 (which is proven immediately after this theorem):

$$E(\|\vec{e}_c\|) \le \sigma\sqrt{n} \quad (40)$$

$\square$.

**Lemma 1:** Assume that we have a function $t(x)$ of a random variable *x*, and a constant $\tau$, such that $t(x) \ge 0$ and $\tau \ge 0$. If $E(t(x)^2) = \tau^2$ then $E(t(x)) \le \tau$.

*Proof:* Consider the variance of $t(x)$:

$$\mathrm{var}(t(x)) = E((t(x) - E(t(x)))^2) \ge 0 \quad (41)$$

We see that:

$$E((t(x) - E(t(x)))^2) = E(t(x)^2 - 2t(x)E(t(x)) + (E(t(x)))^2) \quad (42)$$

$$= E(t(x)^2) - 2\,(E(t(x)))^2 + (E(t(x)))^2 \tag{43}$$

$$= E(t(x)^2) - (E(t(x)))^2 \tag{44}$$

$$= \tau^2 - (E(t(x)))^2 \tag{45}$$

Combining these results, we get:

$$\tau^2 - (E(t(x)))^2 \geq 0 \tag{46}$$

Thus:

$$\tau \geq E(t(x)) \tag{47}$$

□.

Lemma 1 is a well-known result (Fraser, 1976).

It is natural to consider models that minimize the error on the training set:

**Theorem 3:** If:

$$\vec{m}(X|X,\vec{y}_i) = \vec{y}_i \tag{48}$$

Then:

$$E(\|\vec{e}_t\|) = 0 \tag{49}$$

$$E(\|\vec{e}_s\|) = E(\|\vec{e}_c\|) \tag{50}$$

$$E(\|\vec{e}_c\|^2) = 2\sigma^2 n \tag{51}$$

$$E(\|\vec{e}_c\|) \leq \sigma\sqrt{2n} \tag{52}$$

*Proof:* Consider the error on the training set:

$$\|\vec{e}_t\| = \|\vec{m}(X|X,\vec{y}_1) - \vec{y}_1\| = \|\vec{y}_1 - \vec{y}_1\| = 0 \tag{53}$$

$$E(\|\vec{e}_t\|) = 0 \tag{54}$$

Let us introduce a new term:

$$\vec{e}_\omega = \vec{m}(X|X,\vec{y}_2) - \vec{y}_1 \tag{55}$$

We see that:

$$\vec{e}_\omega = \vec{m}(X|X,\vec{y}_2) - \vec{m}(X|X,\vec{y}_1) = \vec{e}_s \tag{56}$$

Therefore:

$$E(\|\vec{e}_s\|) = E(\|\vec{e}_\omega\|) = E(\|\vec{e}_c\|) \tag{57}$$

Now, consider the cross-validation error:

$$\vec{e}_c = \vec{m}(X|\ X, \vec{y}_1) - \vec{y}_2 \ = \ \vec{y}_1 - \vec{y}_2 \tag{58}$$

Recall (14):

$$\begin{aligned} \vec{y}_1 &= \vec{f}(X) + \sigma\vec{z}_1 \\ \vec{y}_2 &= \vec{f}(X) + \sigma\vec{z}_2 \end{aligned} \tag{59}$$

It follows that:

$$\vec{e}_c \ = \ \sigma\vec{z}_1 - \sigma\vec{z}_2 \ = \ \sigma\,(\vec{z}_1 - \vec{z}_2) \tag{60}$$

Let us introduce a new term $\vec{z}_\delta$, where $\vec{z}_\delta \ = \ \vec{z}_1 - \vec{z}_2$. Thus $\vec{e}_c \ = \ \sigma\vec{z}_\delta$. Since $\vec{z}_1$ and $\vec{z}_2$ are sequences of independent samples from a standardized distribution (mean 0, variance 1), it follows that $\vec{z}_\delta$ has mean 0 and variance 2 (Fraser, 1976). Let us introduce another new term $\vec{z}_\nu$, where:

$$\vec{z}_\nu \ = \ \frac{\vec{z}_\delta}{\sqrt{2}} \ = \ \frac{\vec{z}_1 - \vec{z}_2}{\sqrt{2}} \tag{61}$$

We see that $\vec{z}_\nu$ is a sequence of samples from a distribution with mean 0 and variance 1, and that:

$$\vec{e}_c \ = \ \sigma\sqrt{2}\vec{z}_\nu \tag{62}$$

Thus:

$$E(\|\vec{e}_c\|^2) \ = \ E(\|\sigma\sqrt{2}\vec{z}_\nu\|^2) \ = \ 2\sigma^2 E(\|\vec{z}_\nu\|^2) \ = \ 2\sigma^2 E\left(\sum_{i=1}^{n} z_{\nu,\,i}^2\right) \ = \ 2\sigma^2 n \tag{63}$$

Finally, applying Lemma 1:

$$E(\|\vec{e}_c\|) \le \sigma\sqrt{2n} \tag{64}$$

□ .

Comparing Theorem 2 with Theorem 3, we see that models that minimize the error on the training set give sub-optimal performance. The squared length of the cross-validation error $E(\|\vec{e}_c\|^2)$ for $\vec{m}(X|\ X, \vec{y}_i) \ = \ \vec{y}_i$ is two times larger than the squared length of the cross-validation error $E(\|\vec{e}_c\|^2)$ for $\vec{m}(X|X,\vec{y}_i) = \vec{f}(X)$. [3]

As we shall see in the next two sections, it is possible to theoretically analyze the expected error on the training set $E(\|\vec{e}_t\|)$ and the instability $E(\|\vec{e}_s\|)$ for some modeling techniques. Section 3 gives some results for linear regression and Section 4 gives some results for instance-based learning.

# 3 Linear Regression

With linear regression, the model *m* is a system of linear equations:

$$\vec{m}(X|X,\vec{y}) = X\vec{b} = \sum_{i=1}^{r} \vec{x}_i b_i \quad (65)$$

The coefficients $\vec{b}$ of the equations are tuned to minimize $\|\vec{e}_t\|$ for the data $(X,\vec{y})$ :

$$\vec{b} = \begin{bmatrix} b_1 \\ \dots \\ b_r \end{bmatrix} \quad (66)$$

We can express the models for the data $(X,\vec{y}_1)$ and for the data $(X,\vec{y}_2)$ as follows (Draper and Smith, 1981; Fraser, 1976; Strang, 1976):

$$\begin{aligned} \vec{m}(X|X,\vec{y}_1) &= P\vec{y}_1 \\ \vec{m}(X|X,\vec{y}_2) &= P\vec{y}_2 \end{aligned} \quad (67)$$

The matrix *P* is called the *projection matrix*:

$$P = X(X^{\mathrm{T}}X)^{-1}X^{\mathrm{T}} \quad (68)$$

$P\vec{y}$ is the projection of the vector $\vec{y}$ onto the column space (the space spanned by the column vectors) of the matrix *X*. It is possible to prove that this projection gives the lowest $\|\vec{e}_t\|$ that it is possible to get by setting the coefficients of the linear equation (Draper and Smith, 1981; Fraser, 1976; Strang, 1976).

An interesting issue in linear regression is the selection of the inputs to the black box. Let us imagine that we have an unlimited supply of potential inputs, and we wish to select a subset consisting of *r* of those inputs. Thus *r* is an adjustable parameter, under the control of the modeler. It is not a fixed value, given by the experimental design. This perspective lets us avoid some of the apparent limitations of linear regression. Although our model *m* must be linear, we can introduce an unlimited number of new inputs, by treating various functions of the given inputs as if they were themselves inputs. The parameter *r* is called the *number of terms* in the linear model (Draper and Smith, 1981).

Draper and Smith (1981) describe the process of selecting the inputs as follows:

> Suppose we wish to establish a linear regression equation for a particular response $Y$ in terms of the basic 'independent' or predictor variables $X_1, X_2, \ldots, X_k$. Suppose further that $Z_1, Z_2, \ldots, Z_r$, all functions of one or more of the $X$'s, represent the complete set of variables from which the equation is to be chosen and that this set includes any functions, such as squares, cross products, logarithms, inverses, and powers thought to be desirable and necessary. Two opposed criteria of selecting a resultant equation are usually involved:
>
> (1) To make the equation useful for predictive purposes we should want our model to include as many $Z$'s as possible so that reliable fitted values can be determined.
>
> (2) Because of the costs involved in obtaining information on a large number of $Z$'s and subsequently monitoring them, we should like the equation to include as few $Z$'s as possible.
>
> The compromise between these extremes is what is usually called *selecting the best regression equation.* There is no unique statistical procedure for doing this.

Draper and Smith's variables $Z_1, Z_2, \ldots, Z_r$ correspond to our variables $\vec{x}_1, \vec{x}_2, \ldots, \vec{x}_r$. We may agree with the criteria of Draper and Smith, but disagree with their explanations of why these criteria are desirable. An alternative explanation would be to say:

(1) To minimize $E(\|\vec{e}_t\|)$ — in other words, to maximize accuracy — we want our model to include as many $Z$'s as possible.

(2) To minimize $E(\|\vec{e}_s\|)$ — in other words, to maximize simplicity or stability — we want our model to include as few $Z$'s as possible.

Theorem 1 tells us that, if we minimize the sum $E(\|\vec{e}_t\|) + E(\|\vec{e}_s\|)$, then we can set an upper bound on the expected cross-validation error $E(\|\vec{e}_c\|)$. This suggests that we should balance the conflicting demands of simplicity and accuracy by minimizing the sum of the expected error on the training set and the instability. We may now prove some theorems that reveal the conflict between accuracy and simplicity.

**Theorem 4:** If $r = n$ and the column vectors of $X$ are linearly independent, then:

$$\vec{m}(X| X,\vec{y}_i) = \vec{y}_i \tag{69}$$

*Proof:* Assume that $r = n$ and the column vectors of $X$ are linearly independent. We need to assume that the column vectors of $X$ are linearly independent in order to calculate the inverse $(X^{\mathrm{T}} X)^{-1}$. Since the column vectors are linearly independent, they span the full $n$-dimensional space in which the vector $\vec{y}_i$ lies. Therefore:

$$\vec{m}(X | X, \vec{y}_i) = P\vec{y}_i = \vec{y}_i \tag{70}$$

$\square$.

Note that linear regression requires that $r \le n$. If $r > n$, then the column vectors cannot be linearly independent.

Theorem 4 is a well-known result in linear regression (Draper and Smith, 1981; Fraser, 1976; Strang, 1976). Combining Theorems 3 and 4, if $r = n$ and the column vectors of $X$ are linearly independent, then:

$$E(\|\vec{e}_t\|) = 0 \tag{71}$$

$$E(\|\vec{e}_s\|) = E(\|\vec{e}_c\|) \tag{72}$$

$$E(\|\vec{e}_c\|^2) = 2\sigma^2 n \tag{73}$$

$$E(\|\vec{e}_c\|) \le \sigma\sqrt{2n} \tag{74}$$

**Theorem 5:** If the column vectors in the matrix $X$ are a subset of the column vectors in the matrix $X'$, then the error on the training set with $X$ is greater than or equal to the error on the training set with $X'$:

$$\vec{e}_t = P\vec{y}_1 - \vec{y}_1 \qquad P = X(X^{\mathrm{T}} X)^{-1} X^{\mathrm{T}} \tag{75}$$

$$\vec{e}_{t'} = P'\vec{y}_1 - \vec{y}_1 \qquad P' = X'(X'^{\mathrm{T}} X')^{-1} X'^{\mathrm{T}} \tag{76}$$

$$\|\vec{e}_t\| \ge \|\vec{e}_{t'}\| \tag{77}$$

$$E(\|\vec{e}_t\|) \ge E(\|\vec{e}_{t'}\|) \tag{78}$$

*Proof:* Assume that the column vectors in the matrix $X$ are a subset of the column vectors in the matrix $X'$. The space spanned by the column vectors in $X$ must be a subspace of the space spanned by the column vectors in $X'$. Therefore the projection $P'\vec{y}_1$ must be closer to $\vec{y}_1$ than the projection $P\vec{y}_1$. Thus:

$$\|\vec{e}_t\| \ge \|\vec{e}_{t'}\| \tag{79}$$

It follows that:

$$E(\|\vec{e}_t\|) \geq E(\|\vec{e}_{t'}\|) \tag{80}$$

$\Box$ .

Theorem 5 is also a well-known result in linear regression (Draper and Smith, 1981; Fraser, 1976; Strang, 1976). The implication of Theorems 4 and 5 is that increasing *r* will decrease the error on the training set. That is, increasing *r* will increase accuracy.

The next theorem shows that increasing *r* will increase instability (as measured by $E(\|\vec{e}_s\|^2)$, not $E(\|\vec{e}_s\|)$). [3] That is, increasing *r* will decrease simplicity.

**Theorem 6:** If *X* has *r* linearly independent column vectors, then:

$$E(\|\vec{e}_s\|^2) = 2\sigma^2 r \tag{81}$$

$$E(\|\vec{e}_s\|) \leq \sigma\sqrt{2r} \tag{82}$$

*Proof:* Assume that *X* has *r* column vectors and they are linearly independent. We need to assume that the column vectors of *X* are linearly independent in order to calculate the inverse $(X^{\mathrm{T}}X)^{-1}$. By the definition of $\vec{e}_s$, we have:

$$\vec{e}_s = P\vec{y}_2 - P\vec{y}_1 = P(\vec{y}_2 - \vec{y}_1) \tag{83}$$

Recall (14):

$$\begin{aligned} \vec{y}_1 &= \vec{f}(X) + \sigma\vec{z}_1 \\ \vec{y}_2 &= \vec{f}(X) + \sigma\vec{z}_2 \end{aligned} \tag{84}$$

It follows that:

$$\vec{e}_s = P(\sigma\vec{z}_2 - \sigma\vec{z}_1) = \sigma P(\vec{z}_2 - \vec{z}_1) \tag{85}$$

Let us introduce a new term $\vec{z}_\nu$, where:

$$\vec{z}_\nu = \frac{\vec{z}_2 - \vec{z}_1}{\sqrt{2}} \tag{86}$$

We see that $\vec{z}_\nu$ is a sequence of samples from a distribution with mean 0 and variance 1, and that:

$$\vec{e}_s = \sigma\sqrt{2}P\vec{z}_\nu \tag{87}$$

Thus:

$$E(\|\vec{e}_s\|) = E(\|\sigma\sqrt{2}P\vec{z}_\nu\|) = \sigma\sqrt{2}E(\|P\vec{z}_\nu\|) \tag{88}$$

Using the Gram-Schmidt procedure, put the column vectors of $X$ into orthonormal form $W$ (Strang, 1976). Since $X$ has $r$ linearly independent column vectors, $W$ has $r$ orthonormal column vectors:

$$W = \begin{bmatrix} \vec{w}_1 & \dots & \vec{w}_r \end{bmatrix} \qquad \vec{w}_j = \begin{bmatrix} w_{1,j} \\ \dots \\ w_{n,j} \end{bmatrix} \tag{89}$$

The following is a useful property of $W$ (Fraser, 1976):

$$W^{\mathrm{T}} W = I \tag{90}$$

We can express the projection matrix $P$ in terms of $W$:

$$P = X(X^{\mathrm{T}}X)^{-1}X^{\mathrm{T}} = WW^{\mathrm{T}} \tag{91}$$

$P$ has the following properties (Strang, 1976):

$$P = P^2 = P^{\mathrm{T}} \tag{92}$$

Consider the term $P\vec{z}_\nu$:

$$P\vec{z}_\nu = WW^{\mathrm{T}}\vec{z}_\nu = \sum_{i=1}^{r} \vec{w}_i \vec{w}_i^{\mathrm{T}} \vec{z}_\nu \tag{93}$$

Now consider the squared length of this term:

$$(P\vec{z}_\nu)^{\mathrm{T}}(P\vec{z}_\nu) = \vec{z}_\nu^{\mathrm{T}} P^{\mathrm{T}} P \vec{z}_\nu = \vec{z}_\nu^{\mathrm{T}} P \vec{z}_\nu = \sum_{i=1}^{r} \vec{z}_\nu^{\mathrm{T}} \vec{w}_i \vec{w}_i^{\mathrm{T}} \vec{z}_\nu \tag{94}$$

$$= \sum_{i=1}^{r} (\vec{w}_i^{\mathrm{T}} \vec{z}_\nu)^{\mathrm{T}} (\vec{w}_i^{\mathrm{T}} \vec{z}_\nu) = \sum_{i=1}^{r} \left( \sum_{j=1}^{n} w_{j,i} z_{\nu,j} \right)^2 \tag{95}$$

Hold $i$ fixed for a moment and consider the squared term:

$$\left( \sum_{j=1}^{n} w_{j,i} z_{\nu,j} \right)^2 = \sum_{j=1}^{n} \sum_{k=1}^{n} (w_{j,i} z_{\nu,j} w_{k,i} z_{\nu,k}) \tag{96}$$

We can represent this as two sums:

$$\sum_{j=1}^{n}\sum_{k=1}^{n}(w_{j,i}z_{\nu,j}w_{k,i}z_{\nu,k}) = \sum_{j=1}^{n}(w_{j,i}z_{\nu,j})^2+\sum_{j=1}^{n-1}\sum_{k=j+1}^{n}2(w_{j,i}z_{\nu,j}w_{k,i}z_{\nu,k}) \tag{97}$$

Let us consider the expected value of the first of the two sums:

$$E\left(\sum_{j=1}^{n}(w_{j,i}z_{\nu,j})^2\right) = \sum_{j=1}^{n}E((w_{j,i}z_{\nu,j})^2) \tag{98}$$

$$= \sum_{j=1}^{n}w_{j,i}^2E(z_{\nu,j}^2) = \sum_{j=1}^{n}w_{j,i}^2 = 1 \tag{99}$$

This follows from the fact that $\vec{w}_i$ is a normalized vector (it is of length 1) and $z_{\nu,j}$ is a sample from a standardized distribution (it is of variance 1). Now, consider the expected value of the second of the two sums:

$$E\left(\sum_{j=1}^{n-1}\sum_{k=j+1}^{n}2(w_{j,i}z_{\nu,j}w_{k,i}z_{\nu,k})\right) = \sum_{j=1}^{n-1}\sum_{k=j+1}^{n}E(2(w_{j,i}z_{\nu,j}w_{k,i}z_{\nu,k})) \tag{100}$$

$$= \sum_{j=1}^{n-1}\sum_{k=j+1}^{n}2w_{j,i}w_{k,i}E(z_{\nu,j}z_{\nu,k}) = 0 \tag{101}$$

This follows from the fact that $j$ does not equal $k$ at any time, and that $z_{\nu,j}$ and $z_{\nu,k}$ are independent samples from a standardized distribution. [4] Thus, we now have:

$$E((P\vec{z}_\nu)^{\mathrm{T}}(P\vec{z}_\nu)) = \sum_{i=1}^{r}1 = r \tag{102}$$

Therefore:

$$E(\|P\vec{z}_\nu\|^2) = r \tag{103}$$

Thus:

$$E(\|\vec{e}_s\|^2) = 2\sigma^2E(\|P\vec{z}_\nu\|^2) = 2\sigma^2r \tag{104}$$

Applying Lemma 1, we get:

$$E(\|\vec{e}_s\|) \le \sigma\sqrt{2r} \tag{105}$$

□ .

Theorem 6 is a variation on a theorem that first appeared in (Turney, 1990).

It is common to view $r$ as a measure of complexity (Fraser, 1976; Turney, 1990). In other words, we can make our model $m$ simpler by decreasing $r$, the number of independent variables in the model. It has been argued elsewhere (Turney, 1990) that Theorem 6 is a justification of our intuitive desire for simplicity. We want to minimize $r$, because decreasing $r$ will decrease the instability $E(\|\vec{e}_s\|)$. Theorem 1 shows why we want to decrease instability.

On the other hand, it is common to view the error on the training set $E(\|\vec{e}_t\|)$ as a measure of accuracy. In other words, we can make our model $m$ more accurate by increasing $r$. Theorems 4 and 5 show that increasing $r$ will increase accuracy.

We see that there is a conflict between simplicity (as measured by $E(\|\vec{e}_s\|)$) and accuracy (as measured by $E(\|\vec{e}_t\|)$). If our goal is to minimize cross-validation error (as measured by $E(\|\vec{e}_c\|)$), then we must find a balance in the conflict between accuracy and simplicity. Theorem 1 shows us how to find the right balance, assuming our goal is to minimize cross-validation error.

For linear regression, it is possible to prove a stronger version of Theorem 1:

**Theorem 7:** For linear regression:

$$E(\|\vec{e}_c\|^2) = E(\|\vec{e}_t\|^2) + E(\|\vec{e}_s\|^2) \tag{106}$$

*Proof:* Let us introduce a new term:

$$\vec{e}_\omega = \vec{m}(X|X,\vec{y}_2) - \vec{y}_1 = P\vec{y}_2 - \vec{y}_1 \tag{107}$$

Consider $\vec{e}_t$ and $\vec{e}_s$:

$$\vec{e}_t = P\vec{y}_1 - \vec{y}_1 \tag{108}$$

$$\vec{e}_s = P\vec{y}_2 - P\vec{y}_1 = P(\vec{y}_2 - \vec{y}_1) \tag{109}$$

$P\vec{y}_i$ is the projection of the vector $\vec{y}_i$ on to the column space of $X$. In linear regression, $\vec{e}_t$ is called the *residual* (Draper and Smith, 1981). It is well-known that the residual is orthogonal to the column space of $X$ (Strang, 1976). It is clear that $\vec{e}_s$ lies in the column space of $X$. Therefore $\vec{e}_t$ and $\vec{e}_s$ are orthogonal. Note that:

$$\vec{e}_\omega = \vec{e}_t + \vec{e}_s \tag{110}$$

By the Pythagorean Theorem:

$$\|\vec{e}_\omega\|^2 = \|\vec{e}_t\|^2 + \|\vec{e}_s\|^2 \quad (111)$$

By the symmetry of the training and testing sets:

$$E(\|\vec{e}_c\|^2) = E(\|\vec{e}_\omega\|^2) = E(\|\vec{e}_t\|^2 + \|\vec{e}_s\|^2) = E(\|\vec{e}_t\|^2) + E(\|\vec{e}_s\|^2) \quad (112)$$

□ .

This leads to another interesting result:

**Theorem 8:** If $X$ has $r$ linearly independent column vectors, then:

$$E(\|\vec{e}_c\|^2) \geq 2\sigma^2 r \quad (113)$$

*Proof:* By Theorem 6:

$$E(\|\vec{e}_s\|^2) = 2\sigma^2 r \quad (114)$$

By Theorem 7:

$$E(\|\vec{e}_c\|^2) = E(\|\vec{e}_t\|^2) + E(\|\vec{e}_s\|^2) \quad (115)$$

Therefore:

$$E(\|\vec{e}_c\|^2) \geq 2\sigma^2 r \quad (116)$$

□ .

Theorem 8 provides a clear justification for minimizing $r$. Of course, we must consider the effect of $r$ on $\vec{e}_t$. That is, we must find a balance between the conflicting demands of simplicity and accuracy. Theorems 1 and 8 show us how to balance these demands.

Note that we have made minimal assumptions about $f$ and $z$. We do not need to assume that $f$ is linear and we do not need to assume that $z$ is normal. We have assumed only that $f$ is a deterministic function and that $\vec{z}$ is a sequence of independent random samples from a standardized probability distribution.

# 4 Instance-Based Learning

With instance-based learning, the model $m(\vec{v} | X, \vec{y})$ is constructed by simply storing the data $(X, \vec{y})$. These stored data are the *instances*. In order to make a prediction for the input $\vec{v}$, we examine the row vectors $\vec{v}_1, \ldots, \vec{v}_n$ of the matrix $X$. In the simplest version of instance-based learning, we look for the row vector $\vec{v}_i$ that is most similar to the input $\vec{v}$. The prediction for the output is $m(\vec{v} | X, \vec{y}) = y_i$, the element of $\vec{y}$ that corresponds to the row vector $\vec{v}_i$ (Kibler *et al.*, 1989). This describes instance-based learning for predicting real-valued attributes. It is easy to see how the same procedure could be used for predicting discrete, symbolic attributes (Aha *et al.*, 1991). Instance-based learning is a form of nearest neighbor pattern recognition (Dasarathy, 1991).

There are many ways that one might choose to measure the similarity between two vectors. Let us assume only that we are using a reasonable measure of similarity. Let us say that a similarity measure $\mathrm{sim}(\vec{u}_1, \vec{u}_2)$ is *reasonable* if:

$$\vec{u}_1 \neq \vec{u}_2 \rightarrow \mathrm{sim}(\vec{u}_1, \vec{u}_2) < \mathrm{sim}(\vec{u}_1, \vec{u}_1) \tag{117}$$

That is, the similarity between distinct vectors is always less than the similarity between identical vectors.

The following will present several different forms of instance-based learning. In order to distinguish these forms, let us use different subscripts $m_1, m_2, \ldots$ for each different version.

Let us start by examining the simple form of instance-based learning described above. We may label this $m_1$. If $\vec{v}_i$ is most similar, of all the row vectors $\vec{v}_1, \ldots, \vec{v}_n$ in the matrix $X$, to the input $\vec{v}$, then the prediction for the output is $m_1(\vec{v} | X, \vec{y}) = y_i$. Let us assume $m_1$ uses a reasonable measure of similarity.

**Theorem 9:** For $m_1$, if no two rows in $X$ are identical, then:

$$\vec{m}_1(X | X, \vec{y}_i) = \vec{y}_i \tag{118}$$

*Proof:* Since $m_1$ uses a reasonable measure of similarity and no two rows in $X$ are identical, it follows that:

$$m_1(\vec{v}_j | X, \vec{y}_i) = y_{i,j} \tag{119}$$

Therefore:

$$\vec{m}_1(X | X, \vec{y}_i) = \vec{y}_i \tag{120}$$

$\square$ .

Combining Theorems 3 and 9, for $m_1$:

$$E(\|\vec{e}_t\|) = 0 \tag{121}$$

$$E(\|\vec{e}_s\|) = E(\|\vec{e}_c\|) \tag{122}$$

$$E(\|\vec{e}_c\|^2) = 2\sigma^2 n \tag{123}$$

$$E(\|\vec{e}_c\|) \le \sigma\sqrt{2n} \tag{124}$$

Theorem 9 shows that the simple form of instance-based learning used by $m_1$ achieves perfect accuracy on the training set. Theorem 9 for instance-based learning is comparable to Theorem 4 for linear regression. We see that $m_1$ is similar to linear regression when $r = n$. We know from Theorems 2 and 3 that $m_1$ is sub-optimal.

Kibler *et al*. (1989) note that $m_1$ may be sensitive to noise, and they suggest a solution, which we may label $m_2$. Given an input vector $\vec{v}$, find the *k* row vectors in *X* that are most similar to $\vec{v}$, where $1 \le k \le n$. The predicted output $m_2(\vec{v} | X, \vec{y})$ is a weighted average of the *k* elements in $\vec{y}$ that correspond to the *k* most similar rows in *X*. The weights are proportional to the degree of similarity between the input vector $\vec{v}$ and each of the *k* rows.

Let $\vec{v}_1, \ldots, \vec{v}_k$ be the *k* row vectors in *X* that are most similar to the input vector $\vec{v}$. Let $w_1, \ldots, w_k$ be weights such that:

$$w_i \propto \mathrm{sim}(\vec{v}, \vec{v}_i) \qquad 0 \le w_i \le 1 \qquad \sum_{i=1}^{k} w_i = 1 \tag{125}$$

Let $y_1, \ldots, y_k$ be the *k* elements in $\vec{y}$ that correspond to the *k* row vectors $\vec{v}_1, \ldots, \vec{v}_k$. The predicted output $m_2(\vec{v} | X, \vec{y})$ is:

$$m_2(\vec{v} | X, \vec{y}) = \sum_{i=1}^{k} w_i y_i \tag{126}$$

Clearly $m_2$ equals $m_1$ when $k = 1$. The parameter *k* is called the *size of the neighborhood* in nearest neighbor pattern recognition (Dasarathy, 1991).

**Theorem 10:** For $m_2$, if no two rows in *X* are identical, then the least stable situation arises when there is a weight $w_i$, such that:

$$w_i = 1 \qquad j \ne i \to w_j = 0 \tag{127}$$

When this situation arises for every row in *X*, we have:

$$E(\|\vec{e}_t\|) = 0 \tag{128}$$

$$E(\|\vec{e}_s\|) = E(\|\vec{e}_c\|) \tag{129}$$

$$E(\|\vec{e}_c\|^2) = 2\sigma^2 n \tag{130}$$

$$E(\|\vec{e}_c\|) \le \sigma\sqrt{2n} \tag{131}$$

The most stable situation arises when:

$$w_i = \frac{1}{k} \qquad i = 1, \ldots, k \tag{132}$$

When this situation arises for every row in $X$, we have:

$$E(\|\vec{e}_s\|^2) = 2\sigma^2 n/k \tag{133}$$

$$E(\|\vec{e}_s\|) \le \sigma\sqrt{2n/k} \tag{134}$$

In general:

$$2\sigma^2 n/k \le E(\|\vec{e}_s\|^2) \le 2\sigma^2 n \tag{135}$$

*Proof:* Let us consider, as an input vector, a single row $\vec{v}_i$ in $X$. Let $\vec{v}_1, \ldots, \vec{v}_k$ be the $k$ row vectors in $X$ that are most similar to the input vector $\vec{v}_i$. Let $w_1, \ldots, w_k$ be weights such that:

$$w_j \propto \mathrm{sim}(\vec{v}_i, \vec{v}_j) \qquad 0 \le w_j \le 1 \qquad \sum_{j=1}^{k} w_j = 1 \tag{136}$$

Let $y_{1,1}, \ldots, y_{1,k}$ be the $k$ elements in $\vec{y}_1$ that correspond to the $k$ row vectors $\vec{v}_1, \ldots, \vec{v}_k$. The predicted output $m_2(\vec{v}_i | X, \vec{y}_1)$ is:

$$m_2(\vec{v}_i | X, \vec{y}_1) = \sum_{j=1}^{k} w_j y_{1,j} \tag{137}$$

Let $y_{2,1}, \ldots, y_{2,k}$ be the $k$ elements in $\vec{y}_2$ that correspond to the $k$ row vectors $\vec{v}_1, \ldots, \vec{v}_k$. The predicted output $m_2(\vec{v}_i | X, \vec{y}_2)$ is:

$$m_2(\vec{v}_i | X, \vec{y}_2) = \sum_{j=1}^{k} w_j y_{2,j} \tag{138}$$

The weights $w_1, \ldots, w_k$ are the same for both $m_2(\vec{v}_i | X, \vec{y}_1)$ and $m_2(\vec{v}_i | X, \vec{y}_2)$, since the weights are based on $X$ alone. We see that: [4]

$$E((m_2(\vec{v}_i | X, \vec{y}_2) - m_2(\vec{v}_i | X, \vec{y}_1))^2) = E\left(\left(\sum_{j=1}^{k} w_j y_{2,j} - \sum_{j=1}^{k} w_j y_{1,j}\right)^2\right) \tag{139}$$

$$= E\left(\left(\sum_{j=1}^{k} \sigma w_j (z_{2,j} - z_{1,j})\right)^2\right) \tag{140}$$

Let us introduce a new term $\vec{z}_\nu$, where:

$$\vec{z}_\nu = \frac{\vec{z}_2 - \vec{z}_1}{\sqrt{2}} \tag{141}$$

Thus:

$$E((m_2(\vec{v}_i | X, \vec{y}_2) - m_2(\vec{v}_i | X, \vec{y}_1))^2) = 2\sigma^2 E\left(\left(\sum_{j=1}^{k} w_j z_{\nu,j}\right)^2\right) \tag{142}$$

Since the random samples are independent and standardized, it follows that:

$$E((m_2(\vec{v}_i | X, \vec{y}_2) - m_2(\vec{v}_i | X, \vec{y}_1))^2) = 2\sigma^2 \sum_{j=1}^{k} E((w_j z_{\nu,j})^2) \tag{143}$$

$$= 2\sigma^2 \sum_{j=1}^{k} w_j^2 E(z_{\nu,j}^2) = 2\sigma^2 \sum_{j=1}^{k} w_j^2 \tag{144}$$

From the constraints on the weights $w_1, \ldots, w_k$, it follows that:

$$\frac{1}{k} \le \sum_{j=1}^{k} w_j^2 \le 1 \tag{145}$$

We reach the upper bound 1 when there is a weight $w_i$, such that:

$$w_i = 1 \qquad j \ne i \to w_j = 0 \tag{146}$$

Since we have a reasonable measure of similarity, the weights are proportional to the similarity, and no two rows in $X$ are identical, it follows that the single non-zero weight can only be the weight $w_i$ corresponding to the row $\vec{v}_i$. We reach the lower bound $1/k$ when:

$$w_j = \frac{1}{k} \qquad j = 1, \ldots, k \tag{147}$$

Suppose the first situation holds, so we are at the upper bound:

$$E((m_2(\vec{v}_i|X,\vec{y}_2) - m_2(\vec{v}_i|X,\vec{y}_1))^2) = 2\sigma^2 \sum_{j=1}^{k} w_j^2 = 2\sigma^2 \tag{148}$$

Suppose the first situation holds for every row vector $\vec{v}_1, \ldots, \vec{v}_n$ in $X$:

$$E(\|\vec{e}_s\|^2) = nE((m_2(\vec{v}_i|X,\vec{y}_2) - m_2(\vec{v}_i|X,\vec{y}_1))^2) = 2n\sigma^2 \tag{149}$$

By Lemma 1:

$$E(\|\vec{e}_s\|) \le \sigma\sqrt{2n} \tag{150}$$

We see that this is the least stable situation. In this case, $m_2$ is essentially the same as $m_1$, so we can apply Theorems 3 and 9:

$$E(\|\vec{e}_t\|) = 0 \tag{151}$$

$$E(\|\vec{e}_s\|) = E(\|\vec{e}_c\|) \tag{152}$$

$$E(\|\vec{e}_c\|^2) = 2\sigma^2 n \tag{153}$$

$$E(\|\vec{e}_c\|) \le \sigma\sqrt{2n} \tag{154}$$

Suppose the second situation holds, so we are at the lower bound:

$$E((m_2(\vec{v}_i|X,\vec{y}_2) - m_2(\vec{v}_i|X,\vec{y}_1))^2) = 2\sigma^2 \sum_{j=1}^{k} w_j^2 = \frac{2\sigma^2}{k} \tag{155}$$

Suppose the second situation holds for every row vector $\vec{v}_1, \ldots, \vec{v}_n$ in $X$:

$$E(\|\vec{e}_s\|^2) = nE((m_2(\vec{v}_i|X,\vec{y}_2) - m_2(\vec{v}_i|X,\vec{y}_1))^2) = \frac{2n\sigma^2}{k} \tag{156}$$

By Lemma 1:

$$E(\|\vec{e}_s\|) \le \sigma\sqrt{2n/k} \tag{157}$$

We see that this is the most stable situation $\square$.

The term $n/k$ plays a role in Theorem 10 similar to the role played by $r$ in Theorem 6. To find the best model (in the sense of minimal cross-validation error) is to find the best balance between accuracy and stability. High values of $r$ and $n/k$ (such as $r = n/k = n$) yield high accuracy (such as $E(\|\vec{e}_t\|) = 0$), but low stability (such as $E(\|\vec{e}_s\|) \le \sigma\sqrt{2n}$).

Low values of $r$ and $n/k$ (such as $r = n/k = 1$) tend to have low accuracy (depending on the data — but even the best data cannot realistically give $E(\|\vec{e}_t\|) = 0$), but high stability (such as $E(\|\vec{e}_s\|) \le \sigma\sqrt{2}$).

Kibler *et al*. (1989) prove a theorem concerning the accuracy of $m_1$. They then discuss the problem of noise and introduce $m_2$. However, they were unable to theoretically demonstrate the benefit of $m_2$:

> While these results establish the appropriateness of IBP [instance-based prediction] for noise-free functions, real-world data require attention to noise. The standard method for tolerating noise is to use some form of averaging. In particular, we believe that the following change to the algorithm $[m_1]$ yields a method $[m_2]$ which works for noise in the function value, but the proof eludes us.

Theorem 10 may be seen as the proof that eluded Kibler *et al*. (1989).

Theorem 10 shows that a straight average, where the weights are all set to $1/k$, is the most stable situation. This suggests that the weights should not be proportional to the similarities. However, we must consider both stability and accuracy. A straight average may be less accurate than a weighted average, where the weights are proportional to the similarities. The choice will depend on the data. Also, it seems likely that, in general, the best balance of accuracy and stability will be a value of $k$ that is much smaller than $n$, $k \ll n$. When $k$ is much smaller than $n$, we may expect that the $k$ nearest neighbors to an input vector $\vec{v}$ will have almost the same degree of similarity. Thus the weights will be nearly identical. This suggests that we will usually be closer to the lower bound $E(\|\vec{e}_s\|) \le \sigma\sqrt{2n/k}$.

It is common practice in instance-based learning to reduce the size of the model by eliminating redundant instances (Aha & Kibler, 1989). Let us consider how this practice affects the stability of the model. For simplicity, let us consider a variation on the basic algorithm $m_1$, rather than a variation on the more complex algorithm $m_2$. Let us use $m_3$ to refer to this new algorithm.

The model $m_3(\vec{v}|X,\vec{y})$ consists of storing $X$ and $\vec{y}$, but with redundant rows eliminated from $X$, and their corresponding elements eliminated from $\vec{y}$. The procedure for deciding whether a row is redundant uses $X$, but not $\vec{y}$. As with $m_1$, we predict the output, given the input $\vec{v}$, by looking for the single most similar row vector in $X$. Let us assume a reasonable measure of similarity.

**Theorem 11:** For $m_3$, if no two rows in $X$ are identical, then:

$$E(\|\vec{e}_s\|^2) = 2\sigma^2 n \tag{158}$$

$$E(\|\vec{e}_s\|) \le \sigma\sqrt{2n} \tag{159}$$

That is, the stability of $m_3$ is the same as the stability of $m_1$.

*Proof:* Let us consider the most extreme case, in which our model has eliminated all of the rows of $X$ except one. Suppose this last remaining row is the $i$-th row $\vec{v}_i$. Then we have:

$$m_3(\vec{v}|X,\vec{y}) = y_i \qquad \vec{m}_3(X|X,\vec{y}_1) = \begin{bmatrix} y_{1,i} \\ \dots \\ y_{1,i} \end{bmatrix} \qquad \vec{m}_3(X|X,\vec{y}_2) = \begin{bmatrix} y_{2,i} \\ \dots \\ y_{2,i} \end{bmatrix} \tag{160}$$

Since the procedure for deciding whether a row is redundant uses $X$, but not $\vec{y}$, the last remaining row must be the same row $\vec{v}_i$ for both $\vec{m}_3(X|X,\vec{y}_1)$ and $\vec{m}_3(X|X,\vec{y}_2)$. Now we have:

$$\vec{e}_s = \vec{m}_3(X|X,\vec{y}_2) - \vec{m}_3(X|X,\vec{y}_1) = \begin{bmatrix} y_{2,i} - y_{1,i} \\ \dots \\ y_{2,i} - y_{1,i} \end{bmatrix} = \begin{bmatrix} \sigma(z_{2,i} - z_{1,i}) \\ \dots \\ \sigma(z_{2,i} - z_{1,i}) \end{bmatrix} \tag{161}$$

Let us introduce a new term $\vec{z}_\nu$, where:

$$\vec{z}_\nu = \frac{\vec{z}_2 - \vec{z}_1}{\sqrt{2}} \tag{162}$$

Thus:

$$E(\|\vec{e}_s\|^2) = E(2n\sigma^2 z_{\nu,i}^2) = 2n\sigma^2 E(z_{\nu,i}^2) = 2n\sigma^2 \tag{163}$$

By Lemma 1:

$$E(\|\vec{e}_s\|) \le \sigma\sqrt{2n} \tag{164}$$

This proof readily generalizes to cases where two to $n$ rows remain in the model $\square$.

Theorem 11 tells us that eliminating rows from $X$ does not affect the stability of instance-based learning. This is conditional on the procedure used for eliminating rows. The proof of Theorem 11 requires the assumption that the procedure for deciding whether a row is redundant uses $X$, but not $\vec{y}$. A natural procedure for eliminating rows is to set a threshold on similarity. If any pair of rows are more similar than the threshold permits, then one member of the pair is eliminated. This procedure uses $X$, but not $\vec{y}$, so it satisfies the assumptions of Theorem 11. If $\vec{y}$ is treated as another column vector in $X$ during this reduction procedure, then the conditions of Theorem 11 are not met. However, Theorem 11 may still be approximately true, since $\vec{y}$ would be only one of $r + 1$ columns, so its influence should be negligible, especially when $r$ is large.

Although eliminating rows does not affect stability, it may affect accuracy. That is, although the stability of $m_3$ is the same as the stability of $m_1$, the accuracy of $m_3$ is the not necessarily same as the accuracy of $m_1$. This will depend on the data. The procedure for eliminating rows should ensure that there is a balance between the costs (in computer memory and time) and the benefits (in accuracy) of having many rows.

We may now prove a stronger version of Theorem 1, analogous to Theorem 7 for linear regression:

**Theorem 12:** Let $\theta$ be the angle (in radians) between $\vec{e}_t$ and $\vec{e}_s$. Let the angle $\theta$ be statistically independent of the lengths $\|\vec{e}_t\|$ and $\|\vec{e}_s\|$. If $\theta$ is a random sample from the uniform distribution on the interval $[0, 2\pi]$, then:

$$E(\|\vec{e}_c\|^2) = E(\|\vec{e}_t\|^2) + E(\|\vec{e}_s\|^2) \tag{165}$$

*Proof:* Let us introduce a new term:

$$\vec{e}_\omega = \vec{m}(X| X,\vec{y}_2) - \vec{y}_1 \tag{166}$$

Consider $\vec{e}_t$ and $\vec{e}_s$:

$$\vec{e}_t = \vec{m}(X| X,\vec{y}_1) - \vec{y}_1 \tag{167}$$

$$\vec{e}_s = \vec{m}(X| X,\vec{y}_2) - \vec{m}(X| X,\vec{y}_1) \tag{168}$$

Note that:

$$\vec{e}_\omega = \vec{e}_t + \vec{e}_s \tag{169}$$

Using some trigonometry, we see that:

$$\|\grave{e}_\omega\|^2 = \|\grave{e}_t\|^2 + \|\grave{e}_s\|^2 + 2\|\grave{e}_t\|\|\grave{e}_s\|\cos\theta \tag{170}$$

Since $\theta$ is a random sample from the uniform distribution on the interval $[0, 2\pi]$, the probability density for $\theta$ is $1 / (2\pi)$:

$$\int_0^{2\pi} \frac{1}{2\pi} d\theta = 1 \tag{171}$$

Let us consider the expectation $E(\|\grave{e}_\omega\|^2)$:

$$E(\|\grave{e}_\omega\|^2) = E(\|\grave{e}_t\|^2 + \|\grave{e}_s\|^2 + 2\|\grave{e}_t\|\|\grave{e}_s\|\cos\theta) \tag{172}$$

$$= E(\|\grave{e}_t\|^2) + E(\|\grave{e}_s\|^2) + E(2\|\grave{e}_t\|\|\grave{e}_s\|\cos\theta) \tag{173}$$

Since the angle $\theta$ is statistically independent of the lengths $\|\grave{e}_t\|$ and $\|\grave{e}_s\|$, we have:

$$E(2\|\grave{e}_t\|\|\grave{e}_s\|\cos\theta) = 2E(\|\grave{e}_t\|\|\grave{e}_s\|)E(\cos\theta) \tag{174}$$

Let us examine $E(\cos\theta)$:

$$E(\cos\theta) = \int_0^{2\pi} \frac{\cos\theta}{2\pi} d\theta = 0 \tag{175}$$

Therefore:

$$E(\|\grave{e}_\omega\|^2) = E(\|\grave{e}_t\|^2) + E(\|\grave{e}_s\|^2) \tag{176}$$

Finally, from the symmetry of the training set and the testing set, we have:

$$E(\|\grave{e}_c\|^2) = E(\|\grave{e}_\omega\|^2) = E(\|\grave{e}_t\|^2) + E(\|\grave{e}_s\|^2) \tag{177}$$

□ .

Theorem 12 is intended to apply to $m_2$. It is conjectured that $m_2$ satisfies the assumptions of Theorem 12, but there is not yet a proof for this.

The theorems in this section have required minimal assumptions. The main assumptions were that the noise is standardized (mean 0 and variance 1) and statistically independent, and that the similarity measure is reasonable.

# 5 Application

If we wish to apply this theory of cross-validation error to a real problem, the first question that arises is how can we estimate the accuracy $E(\|\vec{e}_t\|)$ and the stability $E(\|\vec{e}_s\|)$? Estimating the accuracy is fairly straightforward. The actual error on the training set $\|\vec{e}_t\|$ is a reasonable estimate of the expected error on the training set $E(\|\vec{e}_t\|)$. Since, in general, the variance of the error on the training set will be greater than zero, $\mathrm{var}(\|\vec{e}_t\|) > 0$, $\|\vec{e}_t\|$ may happen to be far from $E(\|\vec{e}_t\|)$ by random chance. However, the variance becomes smaller as *n* becomes larger, so $\|\vec{e}_t\|$ is a reasonable estimate of the expected error on the training set $E(\|\vec{e}_t\|)$, especially when *n* is large.

If we are not careful, we may undermine the value of $\|\vec{e}_t\|$ as an estimate of $E(\|\vec{e}_t\|)$. Suppose we use the data $(X, \vec{y}_1)$ to arbitrarily freeze our model:

$$\vec{m}(X| X,\vec{y}_i) = \vec{y}_1 \tag{178}$$

Note that this is quite different from $m_1$ in instance-based learning, where:

$$\vec{m}_1(X| X,\vec{y}_i) = \vec{y}_i \tag{179}$$

In (178), the model is static; it does not respond to changes in the given data $(X, \vec{y}_i)$. In (179), the model is dynamic; it adjusts to the data. In (178), $\|\vec{e}_t\| = 0$, but $\|\vec{e}_t\|$ is not a good estimate of $E(\|\vec{e}_t\|)$. We have destroyed the symmetry (14) between the testing data and the training data, by arbitrarily freezing the model on the training data $(X, \vec{y}_1)$. The equation (178) is making the claim that the value of $\vec{y}_i$ in the data has no influence on the prediction of the model. If this is true, then it is just a statistical fluke that $\|\vec{e}_t\| = 0$ for the data $(X, \vec{y}_1)$. A better estimate of $E(\|\vec{e}_t\|)$ would be the error of the model (178) on the data $(X, \vec{y}_2)$.

Estimating the instability $E(\|\vec{e}_s\|)$ is slightly more difficult. If we are using linear regression or instance-based learning, then we may apply the theorems of Sections 3 and 4. We need only estimate the level of the noise $\sigma$. This may be done either empirically or theoretically. The theoretical approach is to examine the black box. We may arrive at an estimate of $\sigma$ from the design of the experimental set-up. The empirical approach is to apply statistical estimators of $\sigma$ (Fraser, 1976). Of course, some combination of these approaches may be fruitful.

When we are using a modeling technique, such as neural networks, where we do not (yet) have theorems like those of Sections 3 and 4 to guide us, it may be more difficult to estimate $E(\|\vec{e}_s\|)$. One possibility is to employ Monte Carlo techniques. We can simulate testing data $(X, \vec{y}_2)$ by randomly perturbing the training data $(X, \vec{y}_1)$, using random numbers generated with a distribution that matches the assumed distribution of the noise. The actual instability $\|\vec{e}_s\|$, which we measure on the simulated testing data $(X, \vec{y}_2)$ and the real training data $(X, \vec{y}_1)$, may be used to estimate the expected instability $E(\|\vec{e}_s\|)$. We may improve this estimate by averaging the instabilities $\|\vec{e}_s\|$ measured from several sets of simulated testing data.

When we can estimate the accuracy $E(\|\vec{e}_t\|)$ and the stability $E(\|\vec{e}_s\|)$, we can then adjust the parameters of our model to minimize the sum $E(\|\vec{e}_t\|) + E(\|\vec{e}_s\|)$. In the case of linear regression, we can adjust the number of terms *r* in the linear equation. In the case of instance-based learning, we can adjust the size of the neighborhood *k*. Theorem 1 tells us that, if we minimize the sum $E(\|\vec{e}_t\|) + E(\|\vec{e}_s\|)$, then we can set an upper bound on the expected cross-validation error $E(\|\vec{e}_c\|)$.

Kibler *et al*. (1989) report the results of an empirical comparison of linear regression and instance-based learning. The form of instance-based learning that they used was essentially what is called here $m_2$. Let us consider their Experiment 1. In Experiment 1, $n = 209$, $r = 4$ (Ein-Dor and Feldmesser, 1987), and $k = 6$ (Kibler *et al.*, 1989). Therefore $n/k = 34.8$. From Theorems 6 and 10, we see that linear regression will be more stable, for these values of *n*, *r*, and *k*, than instance-based learning. Kibler *et al*. (1989) did not use cross-validation testing, but they did report the error on the training set. They found that linear regression and instance-based learning had very similar training set error. We may take this to indicate that they would have very similar expected training set error. Although we do not know the level of noise $\sigma$, if there is a significant amount of noise, then Theorems 1, 6, and 10, taken together, suggest that linear regression will have lower cross-validation error than instance-based learning. This result must certainly depend on the data. For some data, linear regression will have a lower cross-validation error than instance-based learning. For other data, instance-based learning will have a lower cross-validation error than linear regression.

# 6 An Example

Let us consider a two-dimensional example. Suppose the training set consists of four points ($n = 4$) in $\Re^2$:

$$\vec{x} = \begin{bmatrix} x_1 \\ x_2 \\ x_3 \\ x_4 \end{bmatrix} = \begin{bmatrix} 0.20 \\ 0.35 \\ 0.60 \\ 0.80 \end{bmatrix} \qquad \vec{y} = \begin{bmatrix} y_1 \\ y_2 \\ y_3 \\ y_4 \end{bmatrix} = \begin{bmatrix} 0.15 \\ 0.85 \\ 0.55 \\ 0.75 \end{bmatrix} \tag{180}$$

Each point has the form $(x_i, y_i)$. Our task is to form a model of this data.

Let us first apply linear regression. As discussed in Section 3, we are not limited to the given inputs $\vec{x}$. We may introduce an unlimited number of functions of $\vec{x}$. Let us consider here the natural number powers of $\vec{x}$:

$$\vec{x}_i = \begin{bmatrix} x_1^{i-1} \\ \dots \\ x_4^{i-1} \end{bmatrix} \qquad i = 1, \dots, 4 \tag{181}$$

Let us consider four different linear regression models, $m_{LR1}, m_{LR2}, m_{LR3}, m_{LR4}$. These four different models are based on four different choices for the inputs:

$$X_i = \begin{bmatrix} \vec{x}_1, \dots, \vec{x}_i \end{bmatrix} \qquad i = 1, \dots, 4 \tag{182}$$

For example:

$$X_3 = \begin{bmatrix} \vec{x}_1, \vec{x}_2, \vec{x}_3 \end{bmatrix} = \begin{bmatrix} x_1^0 & x_1^1 & x_1^2 \\ \dots & \dots & \dots \\ x_4^0 & x_4^1 & x_4^2 \end{bmatrix} = \begin{bmatrix} 1 & x_1 & x_1^2 \\ \dots & \dots & \dots \\ 1 & x_4 & x_4^2 \end{bmatrix} \tag{183}$$

Our four models are:

$$P_i = X_i (X_i^{\mathrm{T}} X_i)^{-1} X_i^{\mathrm{T}} \qquad i = 1, \ldots, 4 \tag{184}$$

$$\vec{m}_{LRi}(X_i | X_i, \vec{y}) = P_i \vec{y} \qquad i = 1, \ldots, 4 \tag{185}$$

Each model $m_{LRi}$ is the polynomial equation of order $i - 1$ that best fits the data. That is, linear regression sets the coefficients of the polynomial equations in order to minimize the error on the training set $\| \vec{e}_t \|$. Note that the instability of the model $m_{LRi}$ is proportional to $r = i$, the number of terms in the model. Figure 1 shows the data fitted with the four linear regression models. The boxes in the plot indicate the data points.

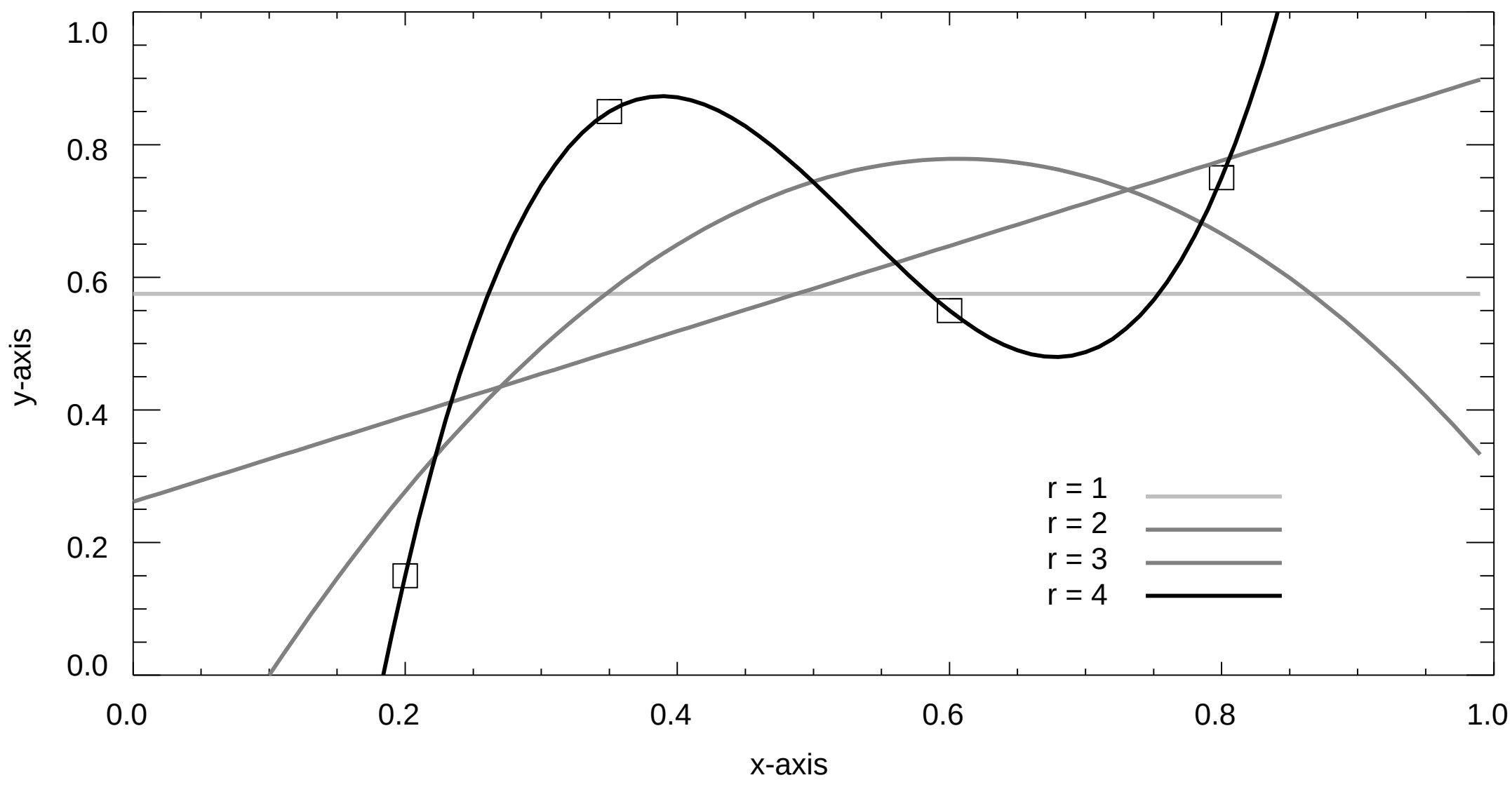

Figure 1. Four points fitted with linear regression.

Let us now apply instance-based learning. Let us consider four different instance-based learning models, $m_{IBL1}$, $m_{IBL2}$, $m_{IBL3}$, $m_{IBL4}$. All four models use the same inputs $X$:

$$X = \vec{x} = \begin{bmatrix} x_1 \\ \ldots \\ x_4 \end{bmatrix} \tag{186}$$

The model $m_{IBLi}$ is essentially the model $m_2$ of Section 4, with $k = i$. The four instance-based learning models use the following measure of similarity (Kibler *et al.*, 1989):

$$\mathrm{sim}(\vec{u}_1, \vec{u}_2) = \sum_j (1 - |u_{1,j} - u_{2,j}|) \qquad (187)$$

Here $|\ldots|$ is the absolute value function. In this example, the rows in $X$ are scalars, not vectors, so the formula for similarity simplifies:

$$\mathrm{sim}(u_1, u_2) = 1 - |u_1 - u_2| \qquad (188)$$

It is clear that this is a reasonable measure of similarity. The four models use a straight average ($w_i = 1/k$) of the $k$ nearest neighbors, in order to get the maximum stability $E(\|\vec{e}_s\|^2) = 2\sigma^2 n/k$. For the model $m_{IBLi}$, we see that the instability is proportional to $n/k = n/i$. Figure 2 shows the data fitted with the four instance-based learning models.

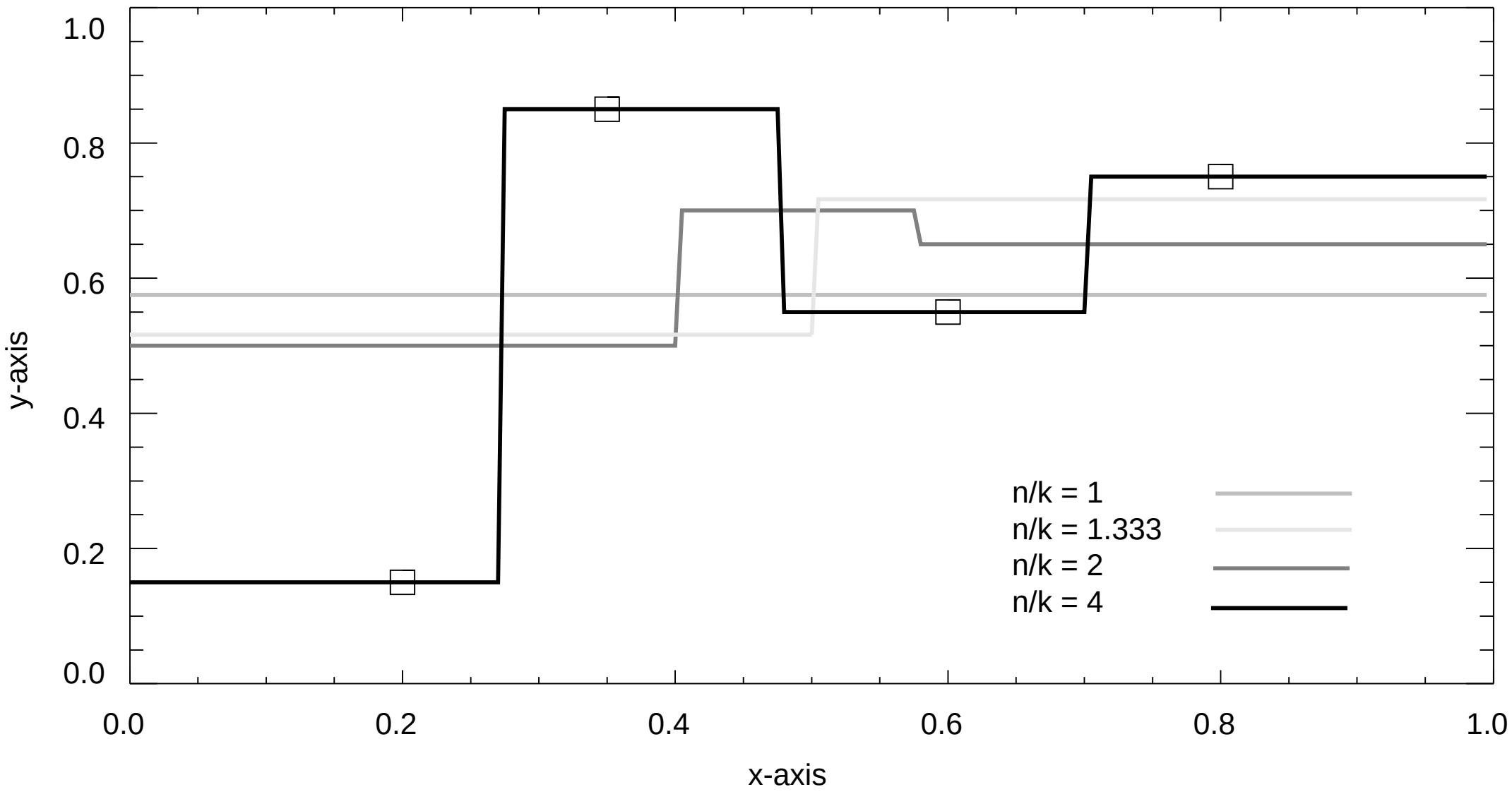

Figure 2. Four points fitted with instance-based learning.

Instability $E(\|\vec{e}_s\|)$ captures an important aspect of our intuitive notion of simplicity. This claim cannot be proven, since it involves intuition. However, examination of Figures 1 and 2 should support the claim. In Figure 1, we see that the curves appear increasingly complex as $r$ increases. From Theorem 6, we know that $E(\|\vec{e}_s\|^2) = 2\sigma^2 r$. Thus instability increases as $r$ increases. In Figure 2, we see that the curves appear increasingly complex as $n/k$ increases. From Theorem 10, we know that $E(\|\vec{e}_s\|^2) = 2\sigma^2 n/k$. Thus instability increases as $n/k$ increases. Figures 1 and 2 show that instability is related to our intuitive feeling of complexity.

Section 5 argued that we can use $\|\hat{e}_t\|$ as an estimate of $E(\|\hat{e}_t\|)$. By the same argument, we can use $\|\hat{e}_t\|^2$ as an estimate of $E(\|\hat{e}_t\|^2)$. For linear regression, we have $E(\|\hat{e}_s\|^2) = 2\sigma^2 r$. For instance-based learning, we have $E(\|\hat{e}_s\|^2) = 2\sigma^2 n/k$, since our weights are a straight average. Given Theorems 7 and 12, we can now estimate $E(\|\hat{e}_c\|^2)$, since $E(\|\hat{e}_c\|^2) = E(\|\hat{e}_t\|^2) + E(\|\hat{e}_s\|^2)$. For linear regression:

$$E(\|\hat{e}_c\|^2) \approx \|\hat{e}_t\|^2 + 2\sigma^2 r \tag{189}$$

Table 1 shows the estimated expected cross-validation error for linear regression, as a function of the level of noise $\sigma^2$. For instance-based learning:

$$E(\|\hat{e}_c\|^2) \approx \|\hat{e}_t\|^2 + 2\sigma^2 n/k \tag{190}$$

Table 2 shows the estimated expected cross-validation error for instance-based learning. If we can estimate the level of noise $\sigma^2$ in our data, then we can use Tables 1 and 2 to find the model with minimal expected cross-validation error $E(\|\hat{e}_c\|^2)$.

| **Model** | **Estimate of** $E(\|\hat{e}_t\|^2)$ | $E(\|\hat{e}_s\|^2)$ | **Estimate of** $E(\|\hat{e}_c\|^2)$ |
|---|---|---|---|
| $m_{LR1}$ | 0.2875 | $2\sigma^2$ | $0.2875 + 2\sigma^2$ |
| $m_{LR2}$ | 0.1999 | $4\sigma^2$ | $0.1999 + 4\sigma^2$ |
| $m_{LR3}$ | 0.1491 | $6\sigma^2$ | $0.1491 + 6\sigma^2$ |
| $m_{LR4}$ | 0.0000 | $8\sigma^2$ | $8\sigma^2$ |

Table 1. Estimated cross-validation error for linear regression.

| **Model** | **Estimate of** $E(\lVert \hat{e}_t \rVert^2)$ | $E(\lVert \hat{e}_s \rVert^2)$ | **Estimate of** $E(\lVert \hat{e}_c \rVert^2)$ |
|---|---|---|---|
| $m_{IBL1}$ | 0.0000 | $8\sigma^2$ | $8\sigma^2$ |
| $m_{IBL2}$ | 0.2650 | $4\sigma^2$ | $0.2650 + 4\sigma^2$ |
| $m_{IBL3}$ | 0.2744 | $2.666\sigma^2$ | $0.2744 + 2.666\sigma^2$ |
| $m_{IBL4}$ | 0.2875 | $2\sigma^2$ | $0.2875 + 2\sigma^2$ |

Table 2. Estimated cross-validation error for instance-based learning.

When $\sigma^2$ is large, we will prefer $m_{LR1}$ and $m_{IBL4}$. These two models have the same expected cross-validation error. When $\sigma^2$ is small, we will prefer $m_{LR4}$ and $m_{IBL1}$. These two models also have the same expected cross-validation error. For intermediate values of $\sigma^2$, it appears from Tables 1 and 2 that linear regression will perform better than instance-based learning, since $m_{LR2}$ has a lower expected cross-validation error than $m_{IBL2}$.

# 7 Related Work

There is some similarity between this theory of cross-validation error and the work in Akaike Information Criterion (AIC) statistics (Akaike, 1970, 1973, 1974; Sakamoto *et al.*, 1986). [5] AIC is a criterion for choosing among competing models, similar to $E(\lVert \hat{e}_c \rVert^2)$.

AIC has the general form (Sakamoto *et al.*, 1986):

$$\text{AIC} = -2l + 2k \tag{191}$$

In this formula, *l* is the maximum log likelihood of the model and *k* is the number of free parameters in the model. This is analogous to the equation in Theorems 7 and 12:

$$E(\lVert \hat{e}_c \rVert^2) = E(\lVert \hat{e}_t \rVert^2) + E(\lVert \hat{e}_s \rVert^2) \tag{192}$$

Both *l* and $E(\lVert \hat{e}_t \rVert^2)$ may be thought of as measures of accuracy, since the maximum log likelihood *l* of the model is closely related to the error on the training set. Both *k* and $E(\lVert \hat{e}_s \rVert^2)$ may be thought of as measures of stability, since Theorem 6 shows that the number of free parameters *k* in the model is closely related to the instability $E(\lVert \hat{e}_s \rVert^2)$.

Just as we want to minimize $E(\|\vec{e}_c\|^2)$, we also want to minimize AIC. The argument for minimizing AIC is that $(-1/2)\,\text{AIC}$ is an asymptotically unbiased estimator of the *mean expected log likelihood*. The mean expected log likelihood is the mean of the expected log likelihood of the maximum likelihood model (Sakamoto *et al.*, 1986). The definition of mean expected log likelihood is somewhat similar to the definition of $E(\|\vec{e}_c\|^2)$. Both definitions involve finding the best fit for one set of outputs $\vec{y}_1$ and then evaluating the fit with a second set of outputs $\vec{y}_2$ (compare (16) and (17) above with (4.8) and (4.9) in (Sakamoto *et al.*, 1986)).

Let us compare AIC and $E(\|\vec{e}_c\|^2)$ for linear regression. For AIC, we have (see (8.29) in (Sakamoto *et al.*, 1986)):

$$l = -\frac{n\log 2\pi\sigma^2}{2} - \frac{\|\vec{e}_t\|^2}{2\sigma^2} \tag{193}$$

$$k = r \tag{194}$$

$$\text{AIC} = -2l + 2k = n\log 2\pi\sigma^2 + \frac{\|\vec{e}_t\|^2}{\sigma^2} + 2r \tag{195}$$

In order to derive this equation for AIC, we must assume that the noise has a normal distribution. For comparison, let us introduce CVC (Cross-Validation Criterion):

$$\text{CVC} = \|\vec{e}_t\|^2 + 2\sigma^2 r \tag{196}$$

We see that CVC is an unbiased estimator of the expected cross-validation error for linear regression:

$$E(\text{CVC}) = E(\|\vec{e}_t\|^2 + 2\sigma^2 r) \tag{197}$$

$$= E(\|\vec{e}_t\|^2) + E(\|\vec{e}_s\|^2) \tag{198}$$

$$= E(\|\vec{e}_c\|^2) \tag{199}$$

In the equations (195) and (196), we may consider $\sigma^2$ and $n$ to be constants, outside of the modeler's control. Therefore AIC and CVC are linearly proportional to each other:

$$c_1 = n\log 2\pi\sigma^2 \tag{200}$$

$$c_2 = 1/\sigma^2 \tag{201}$$

$$\text{AIC} = c_1 + c_2\text{CVC} \tag{202}$$

Thus minimizing AIC is equivalent to minimizing CVC.

There are some differences between the approach in this paper and the AIC approach. Note that the derivation of CVC does not require the assumption that the noise has a normal distribution, unlike the derivation of AIC. Furthermore, CVC is an unbiased estimator of $E(\|\vec{e}_c\|^2)$, but AIC is only an asymptotically unbiased estimator of the mean expected log likelihood. [6] Perhaps the most profound difference between the two approaches is that mean expected log likelihood is defined using the *parameters* (the coefficients $\vec{b}$ of equation (65)) of the model (Sakamoto *et al.*, 1986), while $E(\|\vec{e}_c\|^2)$ is defined without reference to parameters. This makes $E(\|\vec{e}_c\|^2)$ particularly suitable for instance-based learning, which is a type of nonparametric analysis (Dasarathy, 1991). In general, machine learning algorithms tend to be nonparametric (in the statistical sense of "parametric"), thus this theory of cross-validation error is particularly suitable for machine learning.

More generally, the work here is related to the view that selecting a model involves a trade-off between bias and variance (Barron, 1984; Eubank, 1988; Geman *et al.*, 1992; Moody, 1991, 1992). Bias and variance are defined as follows (Geman *et al.*, 1992):

$$\text{bias:} \qquad [E_{X,\vec{y}}(m(\vec{v}|X,\vec{y})) - E(y|\vec{v})]^2 \tag{203}$$

$$\text{variance:} \qquad E_{X,\vec{y}}[(m(\vec{v}|X,\vec{y}) - E_{X,\vec{y}}(m(\vec{v}|X,\vec{y})))^2] \tag{204}$$

The bias of the model $m(\vec{v}|X,\vec{y})$ for the value $\vec{v}$ is the square of the difference between the expected output of the model $E_{X,\vec{y}}(m(\vec{v}|X,\vec{y}))$ and the expected value of $y$, given $\vec{v}$. Note that expected value of $y$ given $\vec{v}$ is $f(\vec{v})$:

$$E(y|\vec{v}) = E(f(\vec{v}) + \sigma z|\vec{v}) = E(f(\vec{v})|\vec{v}) + E(\sigma z|\vec{v}) = f(\vec{v}) \tag{205}$$

The expected output of the model is calculated with respect to $X$ and $\vec{y}$, which requires a probability distribution over $X$ and $\vec{y}$. The variance of the model $m(\vec{v}|X,\vec{y})$ for the value $\vec{v}$ is the expected value of the square of the difference between the actual output of the model and the expected output of the model.

It is possible to prove that the expected value of the square of the difference between the output of the model and the expected value of $y$ given $\vec{v}$ is the sum of the bias and the variance for $\vec{v}$ (Geman *et al.*, 1992):

$$\begin{aligned} E_{X, \vec{y}} [ (m(\vec{v} | X, \vec{y}) - E(y | \vec{v}))^2 ] &= E_{X, \vec{y}} [ (m(\vec{v} | X, \vec{y}) - f(\vec{v}))^2 ] \\ = [E_{X, \vec{y}}(m(\vec{v} | X, \vec{y})) - E(y | \vec{v})]^2 &+ E_{X, \vec{y}} [ (m(\vec{v} | X, \vec{y}) - E_{X, \vec{y}}(m(\vec{v} | X, \vec{y})))^2 ] \end{aligned} \tag{206}$$

This is analogous to the relationship between expected cross-validation error, expected training set error, and expected instability (Theorems 7 and 12):

$$\begin{aligned} E(\|\vec{e}_c\|^2) &= E(\|\vec{m}(X | X, \vec{y}_1) - \vec{y}_2\|^2) = E(\|\vec{e}_t\|^2) + E(\|\vec{e}_s\|^2) \\ &= E(\|\vec{m}(X | X, \vec{y}_1) - \vec{y}_1\|^2) + E(\|\vec{m}(X | X, \vec{y}_2) - \vec{m}(X | X, \vec{y}_1)\|^2) \end{aligned} \tag{207}$$

Expected training set error is related to bias and expected instability is related to variance.

Bias and variance, as defined by (Geman *et al.*, 1992), are difficult to use in practice, for several reasons:

1. To calculate the expectation $E_{X, \vec{y}}$ requires a probability distribution over $X$ and $\vec{y}$. In most applications, the probability distribution will not be known. With $E(\|\vec{e}_t\|^2)$ and $E(\|\vec{e}_s\|^2)$, it is assumed that the actual, observed value of $X$ has a probability of one, and other possible values of $X$ have a probability of zero. It is assumed that $\vec{y}$ has an arbitrary probability distribution with mean $\vec{f}(X)$ and variance $\sigma^2$. When there is no prior information about the probability distribution over $X$ and $\vec{y}$, these assumptions seem to be the weakest, useful, reasonable assumptions.
2. Bias and variance are functions of $\vec{v}$. With $E(\|\vec{e}_t\|^2)$ and $E(\|\vec{e}_s\|^2)$, in effect, $\vec{v}$ ranges over the rows in $X$. This is consistent with the assumption that the actual, observed value of $X$ has a probability of one.
3. Bias and variance cannot be calculated when $E(y | \vec{v})$ is unknown. Bias compares the model $m$ to the underlying deterministic function $f$, while $E(\|\vec{e}_t\|^2)$ compares the model $m$ to the actual, observed value $y$. In essence, $y$ is used as a substitute for the unknown function $f$. Variance compares the actual value of the model $m$ to the mean value of the model $m$, while $E(\|\vec{e}_s\|^2)$ compares the actual value of the model $m$ to another possible value of the model $m$.

In summary, we may think of the expected training set error and the expected instability as practical ways of evaluating bias and variance.

# 8 Future Work

There is another source of error in cross-validation testing, that has not been examined here. This is the error that the model $m$ makes when it is confronted with input values in the testing set that it has not seen in the training set. The weak point of this theory is the assumption (14) that $X$ is the same in the training data and the testing data. The reasons for this assumption were discussed in Section 2. It makes the mathematics simpler and it lets us avoid making any assumptions about $f$ the deterministic aspect of the black box. The main implication of the assumption is that we may be underestimating the cross-validation error. This is an area for future research. It may be possible to prove some interesting results by making some weak assumptions about $f$, such as the assumptions made in Kibler *et al*. (1989). For example, we might assume that there is an upper bound on the absolute value of the derivative of $f$. In other words, the slope of $f$ cannot be too steep.

This paper has dealt exclusively with predicting real-valued attributes. Another paper will deal with the extension of the theory to predicting symbolic attributes.

Another open problem is showing that $m_2$ satisfies the assumptions of Theorem 12. It seems likely that it does, but there is not yet a proof.

# 9 Conclusion

This paper presents three main results concerning error in cross-validation testing of algorithms for predicting real-valued attributes. First, cross-validation error can be thought of as having two components, error on the training set and instability of the model. In other words, a good model should be both accurate and stable. Second, for linear regression, the number of terms in the linear equations affects both accuracy and stability. Increasing the number of terms will improve accuracy. Decreasing the number of terms will improve stability. Third, for instance-based learning, the size of the neighborhood affects both accuracy and stability. Increasing the size of the neighborhood will improve stability. Decreasing the size of the neighborhood will improve accuracy.

# Acknowledgments

I would like to thank Professors Malcolm Forster and Elliott Sober of the University of Wisconsin for discussing their own, related work with me. I would also like to thank an anonymous referee of the *Journal of Experimental and Theoretical Artificial Intelligence* for several helpful comments.

# Notes

1. This point was made by an anonymous referee of the *Journal of Experimental and Theoretical Artificial Intelligence*.
2. Remember that the concern here is prediction, not causal modelling. A causal model might be able to account for the noise after the fact, but our task is to predict the noise before the fact. We assume that the random noise $\vec{z}$ is independent of $X$. The best that we can do is to guess the mean of the noise. This follows from the Cramér-Rao inequality (Fraser, 1976).
3. The theorems here often mention both the expected squared length (e.g. $E(\|\vec{e}_c\|^2)$) and the expected length (e.g. $E(\|\vec{e}_c\|)$). This is because the relation between expected squared length and expected length is not simple, as we can see from Lemma 1. For example, in Theorem 6 we have:

$$E(\|\vec{e}_s\|^2) = 2\sigma^2 r \tag{208}$$

$$E(\|\vec{e}_s\|) \le \sigma\sqrt{2r} \tag{209}$$

   From (208) we might expect to derive $E(\|\vec{e}_s\|) = \sigma\sqrt{2r}$, but Lemma 1 stands in our way. It would have been possible to refer exclusively to expected squared length, but it was felt that the reader may benefit from seeing both expected squared length and expected length.
4. Since $z_{\nu,j}$ and $z_{\nu,k}$ are independent samples from a distribution with a mean of zero, it follows that $E(z_{\nu,j}z_{\nu,k}) = E(z_{\nu,j})E(z_{\nu,k}) = 0$.
5. The relation of this work to Akaike's work was pointed out to me by Malcolm Forster of the University of Wisconsin.
6. It is important to be clear about the distinction between an estimator and the quantity that it estimates. AIC and CVC are both estimators; they can be calculated exactly for a given model and data. AIC is an estimator of the mean expected log likelihood and CVC is an estimator of the expected cross-validation error for linear regression. The mean expected log likelihood and the expected cross-validation error cannot be calculated exactly; they can only be estimated.

# References

Aha, D.W., Kibler, D. (1989) Noise-tolerant instance-based learning algorithms, *Proceedings of the Eleventh International Joint Conference on Artificial Intelligence*, 794-799.

Aha, D.W., Kibler, D., & Albert, M.K. (1991) Instance-based learning algorithms, *Machine Learning*, 6:37-66.

Akaike, H. (1970) Statistical predictor identification, *Annals of the Institute of Statistical Mathematics*, 22:203-217.

Akaike, H. (1973) Information theory and an extension of the maximum likelihood principle, *Second International Symposium on Information Theory*, edited by B.N. Petrov and F. Csaki (Budapest: Akademia Kiado).

Akaike, H. (1974) A new look at the statistical model identification, *IEEE Transactions on Automatic Control*, AC-19: 716-723.

Barron, A.R. (1984) Predicted squared error: a criterion for automatic model selection, in *Self-organizing Methods in Modeling: GMDH Type Algorithms*, edited by S.J. Farlow (New York: Marcel Dekker).

Dasarathy, B.V. (1991) *Nearest Neighbor Pattern Classification Techniques*, Edited collection (California: IEEE Press).

Draper, N.R. & Smith, H. (1981) *Applied Regression Analysis*, Second Edition (New York: John Wiley & Sons).

Ein-Dor, P. & Feldmesser, J. (1987) Attributes of the performance of central processing units: a relative performance prediction model, *Communications of the ACM*, 30:308-317.

Eubank, R.L. (1988) *Spline Smoothing and Nonparametric Regression* (New York: Marcel Dekker).

Fraser, D.A.S. (1976) *Probability and Statistics: Theory and Applications* (Massachusetts: Duxbury Press).

Geman, S., Bienenstock, E., & Doursat, R. (1992) Neural networks and the bias/variance dilemma, *Neural Computation*, 4:1-58.

Kibler, D., Aha, D.W., & Albert, M.K. (1989) Instance-based prediction of real-valued attributes, *Computational Intelligence*, 5:51-57.

Moody, J.E. (1991) Note on generalization, regularization, and architecture selection in nonlinear learning systems, *First IEEE-SP Workshop on Neural Networks for Signal Processing* (California: IEEE Press).

Moody, J.E. (1992) The effective number of parameters: an analysis of generalization and regularization in nonlinear learning systems, in *Advances in Neural Information Processing Systems 4*, edited by J.E. Moody, S.J. Hanson, and R.P. Lippmann (California: Morgan Kaufmann).

Sakamoto, Y., Ishiguro, M., & Kitagawa, G. (1986) *Akaike Information Criterion Statistics* (Dordrecht, Holland: Kluwer Academic Publishers).

Strang, G. (1976) *Linear Algebra and Its Applications* (New York: Academic Press).

Turney, P.D. (1990) The curve fitting problem: a solution, *British Journal for the Philosophy of Science*, 41:509-530.

# Figures

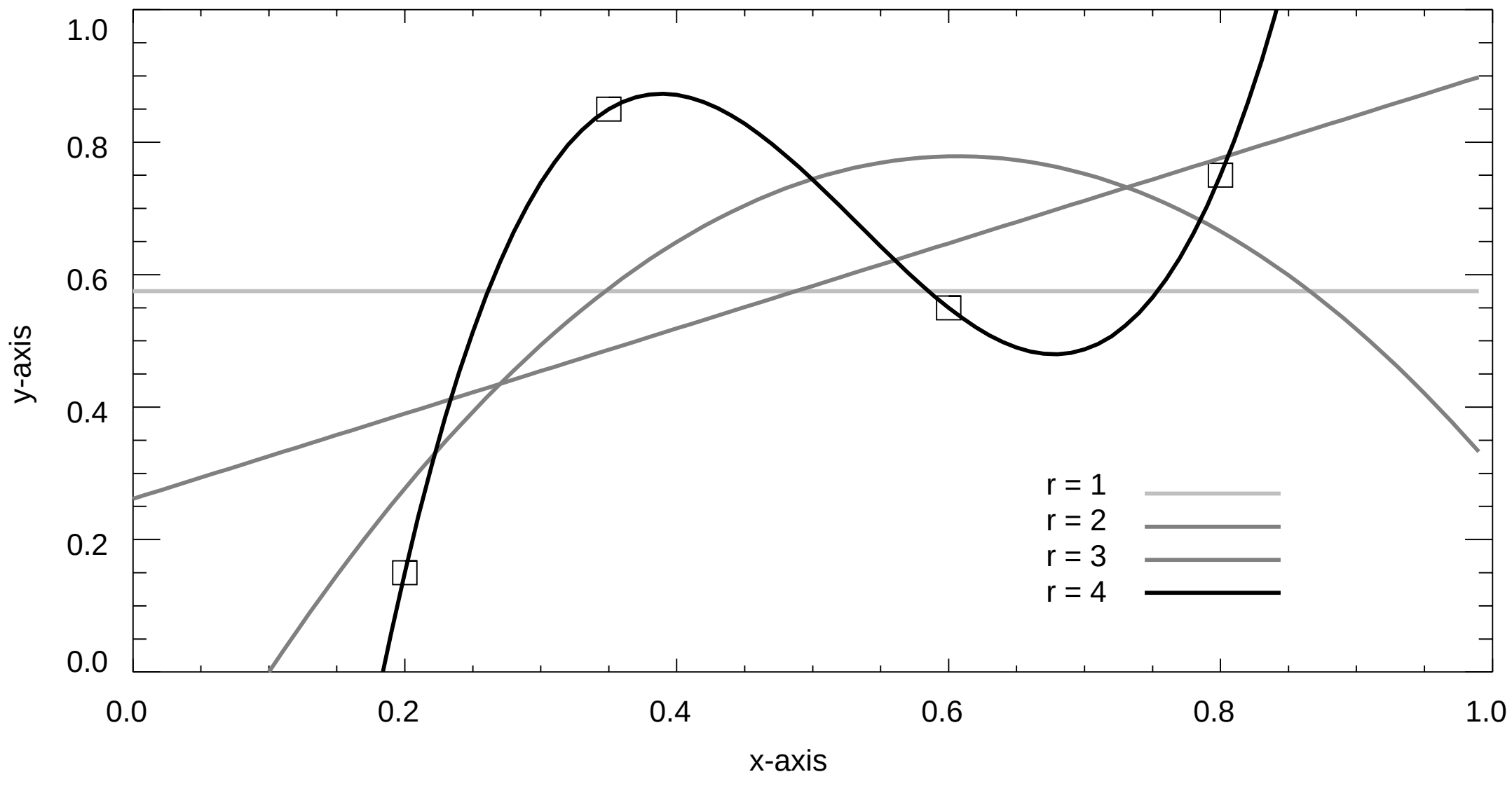

Figure 1. Four points fitted with linear regression.

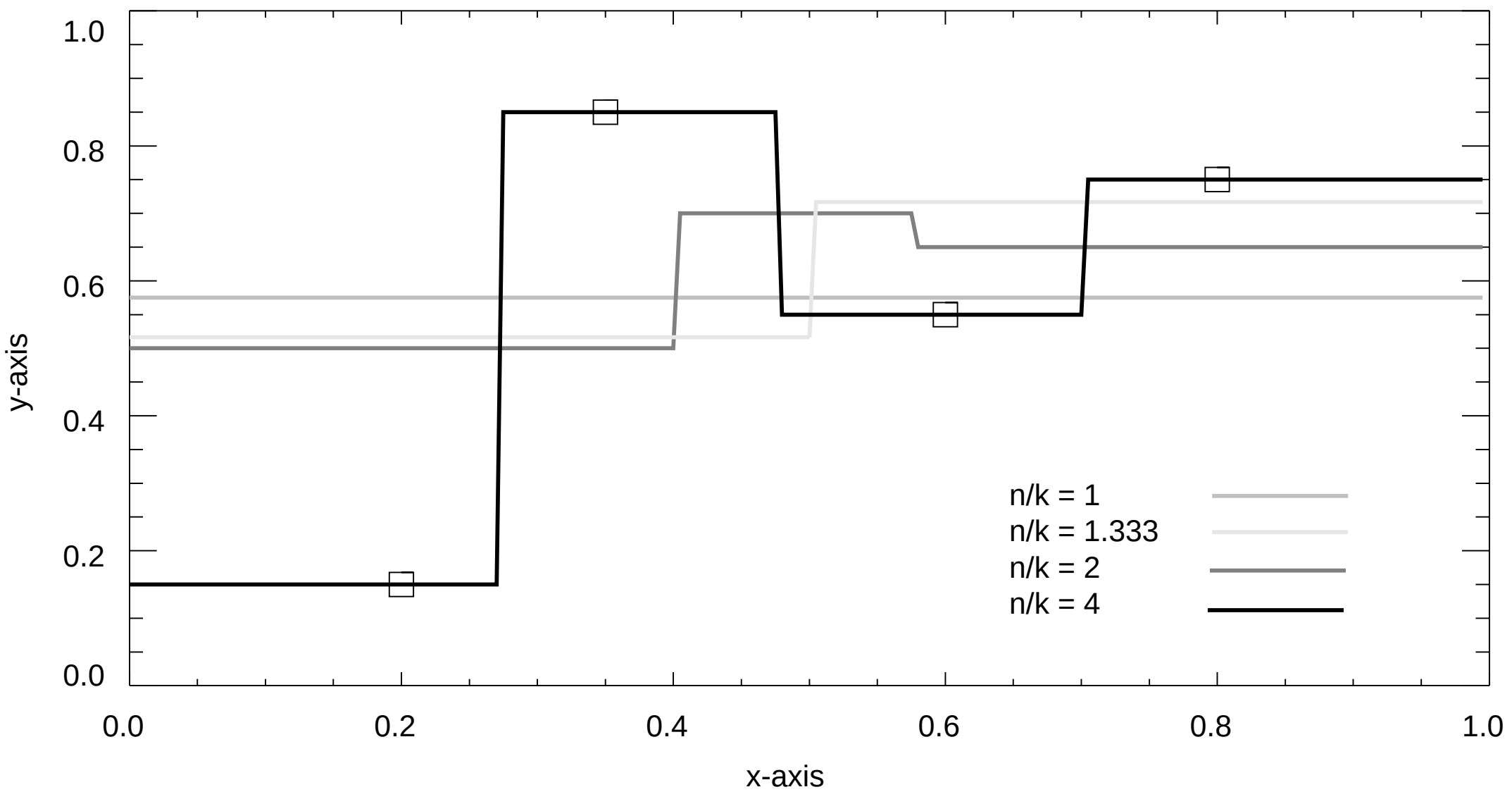

Figure 2. Four points fitted with instance-based learning.

# Tables

| **Model** | **Estimate of** $E(\|\grave{e}_t\|^2)$ | $E(\|\grave{e}_s\|^2)$ | **Estimate of** $E(\|\grave{e}_c\|^2)$ |
|---|---|---|---|
| $m_{LR1}$ | 0.2875 | $2\sigma^2$ | $0.2875 + 2\sigma^2$ |
| $m_{LR2}$ | 0.1999 | $4\sigma^2$ | $0.1999 + 4\sigma^2$ |
| $m_{LR3}$ | 0.1491 | $6\sigma^2$ | $0.1491 + 6\sigma^2$ |
| $m_{LR4}$ | 0.0000 | $8\sigma^2$ | $8\sigma^2$ |

Table 1. Estimated cross-validation error for linear regression.

| **Model** | **Estimate of** $E(\|\vec{e}_t\|^2)$ | $E(\|\vec{e}_s\|^2)$ | **Estimate of** $E(\|\vec{e}_c\|^2)$ |
|---|---|---|---|
| $m_{IBL1}$ | 0.0000 | $8\sigma^2$ | $8\sigma^2$ |
| $m_{IBL2}$ | 0.2650 | $4\sigma^2$ | $0.2650 + 4\sigma^2$ |
| $m_{IBL3}$ | 0.2744 | $2.666\sigma^2$ | $0.2744 + 2.666\sigma^2$ |
| $m_{IBL4}$ | 0.2875 | $2\sigma^2$ | $0.2875 + 2\sigma^2$ |

Table 2. Estimated cross-validation error for instance-based learning.